\documentclass[a4paper,fleqn]{cas-dc}
\usepackage{lineno,hyperref}
\modulolinenumbers[5]

\usepackage{amsmath,amsfonts}
\usepackage{algorithmic}
\usepackage{algorithm}
\usepackage{array}
\usepackage[caption=false,font=normalsize,labelfont=sf,textfont=sf]{subfig}
\usepackage{textcomp}
\usepackage{stfloats}
\usepackage{url}
\usepackage{graphicx}
\usepackage{verbatim}
\usepackage{bm}
\usepackage{cases}
\usepackage{multirow}
\usepackage{multicol}
\usepackage{booktabs}
\usepackage{chngcntr}

\newcommand{\INDSTATE}[1][1]{\STATE\hspace{#1\algorithmicindent}}
\usepackage[authoryear]{natbib}

\def\tsc#1{\csdef{#1}{\textsc{\lowercase{#1}}\xspace}}
\tsc{WGM}
\tsc{QE}

\begin{document}
\let\WriteBookmarks\relax
\def\floatpagepagefraction{1}
\def\textpagefraction{.001}

\shorttitle{ST-DASegNet}    

\shortauthors{Qi Zhao, Shuchang et al.}   

\title [mode = title]{Self-Training Guided Disentangled Adaptation for Cross-Domain Remote Sensing Image Semantic Segmentation}  



%

\author[1]{Qi Zhao}[type=author]



\ead{zhaoqi@buaa.edu.cn}



\affiliation[1]{organization={Beihang University},
            addressline={Xueyuan Road No.37, Haidian district}, 
            city={Beijing},
            postcode={100191}, 
            state={},
            country={China}}

\affiliation[2]{organization={University of Liverpool},
            addressline={Foundation Building,
Brownlow Hill,}, 
            city={Liverpool},
            postcode={L693BX}, 
            state={},
            country={UK}}

\author[1]{Shuchang Lyu}[type=author, orcid=0000-0001-9769-7083]


\ead{lyushuchang@buaa.edu.cn}



\author[1]{Hongbo Zhao}[type=author]
\cormark[1]
\ead{bhzhb@buaa.edu.cn}
\author[1]{Binghao Liu}[type=author]
\ead{liubinghao@buaa.edu.cn}
\author[1]{Lijiang Chen}[type=author]
\ead{chenlijiang@buaa.edu.cn}
\author[2]{Guangliang Cheng}[type=author]
\cormark[2]
\ead{Guangliang.Cheng@liverpool.ac.uk}

\cortext[1]{Primary Corresponding author}
\cortext[2]{Secondary Corresponding author}



\begin{abstract}
Remote sensing (RS) image semantic segmentation using deep convolutional neural networks (DCNNs) has shown great success in various applications. However, the high dependence on annotated data makes it challenging for DCNNs to adapt to different RS scenes. To address this challenge, we propose a cross-domain RS image semantic segmentation task that considers ground sampling distance, remote sensing sensor variation, and different geographical landscapes as the main factors causing domain shifts between source and target images. To mitigate the negative impact of domain shift, we propose a self-training guided disentangled adaptation network (ST-DASegNet) that consists of source and target student backbones to extract source-style and target-style features. To align cross-domain single-style features, we adopt feature-level adversarial learning. We also propose a domain disentangled module (DDM) to extract universal and distinct features from single-domain cross-style features. Finally, we fuse these features and generate predictions using source and target student decoders. Moreover, we employ an exponential moving average (EMA) based cross-domain separated self-training mechanism to ease the instability and disadvantageous effect during adversarial optimization. Our experiments on several prominent RS datasets (Potsdam, Vaihingen, and LoveDA) demonstrate that ST-DASegNet outperforms previous methods and achieves new state-of-the-art results. Visualization and analysis also confirm the interpretability of ST-DASegNet. The code is publicly available at \url{https://github.com/cv516Buaa/ST-DASegNet}.
\end{abstract}


\begin{highlights}
\item We propose a self-training guided disentangled adaptation network (ST-DASegNet) to address the cross-domain RS image semantic segmentation task. Extensive experiments on several predominant datasets (Potsdam, Vaihingen, and LoveDA) demonstrate that ST-DASegNet outperforms previous state-of-the-art methods.
\item We propose a domain disentangled module (DDM) in ST-DASegNet that extracts cross-domain universal features and purifies single-domain distinct features. This provides insight into the use of feature disentangling to bridge the domain gap and improve the adaptability of deep learning models to different RS scenes.
\item We adopt feature-level adversarial learning in ST-DASegNet to align cross-domain single-style features, which enhances the feature consistency of features from cross-domain images.
\item We propose an efficient exponential moving average (EMA) based cross-domain separated self-training paradigm in ST-DASegNet and integrate it with adversarial learning. This design rectifies the representation tendency and stabilizes optimization during adversarial training, leading to improved segmentation performance.
\end{highlights}


\maketitle

\section{Introduction}\label{sec::intro}
\par Remote sensing (RS) technology has been broadly applied in various real-world vision tasks, such as remote sensing scene classification~\cite{AID, NWPU, MGML, LCS}, object detection~\cite{DOTA, HRSC, CRPN-SFNet}, and semantic segmentation~\cite{UAVID, BSNet, SS_att}. Among these applications, remote sensing scene image semantic segmentation has garnered significant research interest, aiming to predict accurate categories for each pixel of RS scene images. Deep Convolutional Neural Networks (DCNNs)~\cite{FCN, PSPNet, DeeplabV3, SegFormer, DLNet} have recently boosted the performance of RS image semantic segmentation tasks. However, their effectiveness heavily relies on large amounts of annotated training samples with similar data distribution to the testing samples. In practical applications, the scene of testing (target) images drastically differ from the scene of training (source) images. To address the domain shift and bridge the domain gap between the source and target images, cross-domain RS image semantic segmentation has become a hot topic.
\par
\begin{figure}[!t]
\centering
\includegraphics[width=1.0\linewidth]{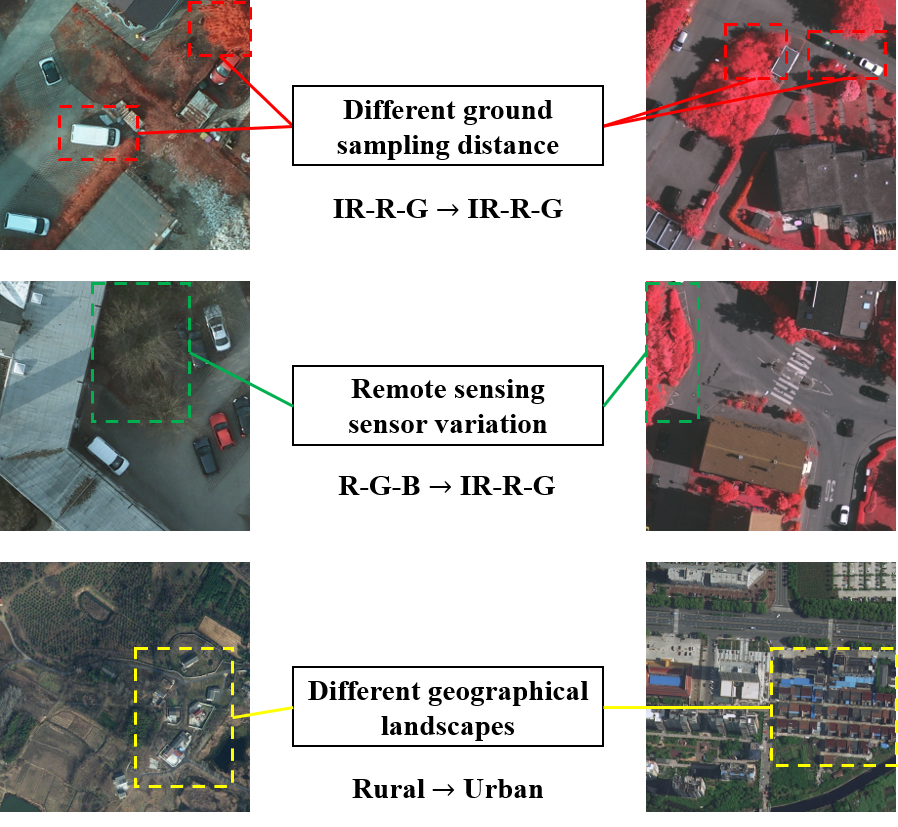}
\caption{Intuitive cases to explain the main problems in cross-domain remote sensing image semantic segmentation.}
\label{Fig1}
\end{figure}

 Cross-domain semantic segmentation task mainly involves training a model with source images and using it to generate pixel-wise predictions on target images, which may have significant domain shifts due to differences between the source and target domains. While many notable methods~\cite{AdapSegNet, CyCADA, cyclegan, UASR, DAFormer} have been proposed to address this problem on natural scene images, the domain shift problem for RS images is primarily caused by differences in ground sampling distance, remote sensing sensor variation, and geographical landscapes, as illustrated in Fig.~\ref{Fig1}. Specifically, different ground sampling distances can result in severe scale variation, remote sensing sensor variation can directly amplify the discrepancy of certain categories between source and target images (e.g., "Tree" on R-G-B and IR-R-G images exhibit different colors), and different geographical landscapes can cause elements of the same category to display different characteristics (e.g., "rural building" may have different patterns than "urban building"). 

\par 
To improve representation generalization and address the domain shift in cross-domain RS image semantic segmentation, we propose four principles as theoretical instruction for our method. First, RS images from different domains have large visual differences, but human beings can still easily distinguish between geographical elements, suggesting that images from different domains contain both unique and invariant features. Second, since RS images cover large geographical areas, image-level alignment using generation techniques~\cite{cyclegan, DA-pix2pix, DA-UDAGAN, DA-DualGAN, DA-ResiDualGAN} (e.g., image translation) may be challenging, while feature-level alignment may be more effective. Thirdly, RS images contain complex geographical information, making the direct adversarial learning process suffer from instability. Finally, while adversarial learning~\cite{AdapSegNet, DA-CLAL, Maetal-MBATA, Wangetal-FGUDA} can fill the domain gap, the lack of target annotations can lead to a representation tendency on source images. Based on these principles, we propose a novel Self-Training Guided Disentangled Adaptation Network (ST-DASegNet) to tackle the cross-domain RS image semantic segmentation task.
\par Following the first two principles, we design a Disentangled Adaptation Network (DASegNet). Specifically, we first design two student backbones to represent source and target images  in two different styles. The source student backbone is designed to represent source and target images as ``source-domain source-style feature'' and ``target-domain source-style feature'', respectively. On the other hand, the target student backbone is designed to represent source and target images as ``source-domain target-style feature'' and ``target-domain target-style feature'', respectively. With this design, we expect to obtain a target-sensitive backbone that is more suitable for target images. Then, towards cross-domain single-style features (source and target output features from the same student backbone), we adopt adversarial learning for feature-level alignment. This can explore the potential of student backbones to adapt source annotations to the target domain. Next, towards single-domain cross-style features (source or target output features from different student backbones), we propose Domain Disentangled Module (DDM) to extract the universal feature and purify distinct features by fusion and disentangling operations. Finally, we design the source student decoder and target student decoder to respectively map source-style and target-style features into final predictions. During optimization with only source-labeled samples, the target student backbone and decoder tend to be more sensitive to target-style features. It means the target student can find the instinct target-style features in both source and target images. Intuitively, the target image contains more instinct target-style features than the source image, so the target student can represent the target image better. Similarly, source student is more suitable for source images.
\par Following the last two principles, we propose an EMA-based cross-domain separated self-training paradigm to further address the representation tendency and optimize the performance of our model. In this paper, we propose two self-training paradigms: ``Decoder-only'' and ``Single-target''. In the ``Decoder-only'' paradigm, we use only two teacher decoders (source teacher decoder and target teacher decoder) instead of the entire teacher network. This efficient paradigm saves a significant amount of training memory and computation cost since the decoder of a segmentation network is much smaller than the backbone.  Additionally, we propose the "Single-target" paradigm, which utilizes the target teacher backbone and target teacher decoder for pseudo label generation. This paradigm provides another perspective on integrating self-training into DASegNet. The EMA-based cross-domain separated self-training paradigm can be seamlessly integrated into DASegNet to form the unified ST-DASegNet.
\par To validate the effectiveness of ST-DASegNet, we conducted extensive experiments on several predominant benchmarks, including ISPRS (Potsdam/Vaihingen)~\cite{ISPRS} and LoveDA~\cite{LoveDA}. Comparison experiments demonstrate that ST-DASegNet outperforms previous SOTA methods and achieves new SOTA performance. Visualization and analysis further illustrate the interpretability of ST-DASegNet. 
\par In summary, the main contributions are listed as follows:
\begin{itemize}
\item We propose a self-training guided disentangled adaptation network (ST-DASegNet) to address the cross-domain RS image semantic segmentation task. Extensive experiments on several predominant datasets (Potsdam, Vaihingen, and LoveDA) demonstrate that ST-DASegNet outperforms previous state-of-the-art methods.
\item We propose a domain disentangled module (DDM) in ST-DASegNet that extracts cross-domain universal features and purifies single-domain distinct features. This provides insight into the use of feature disentangling to bridge the domain gap and improve the adaptability of deep learning models to different RS scenes.
\item We adopt feature-level adversarial learning in ST-DASegNet to align cross-domain single-style features, which enhances the feature consistency of features from cross-domain images.
\item We propose an efficient EMA-based cross-domain separated self-training paradigm in ST-DASegNet and integrate it with adversarial learning. This design rectifies the representation tendency and stabilizes optimization during adversarial training, leading to improved segmentation performance.
\end{itemize}
\section{Related Work} \label{sec::RelatedWork}
\subsection{Semantic Segmentation on RS Images}
\par Semantic segmentation is a classical computer vision task, which plays an important role in many real-world applications. Fully convolutional networks (FCN)~\cite{FCN} first proposes an end-to-end deep learning architecture for semantic segmentation. From then, many notable generalized segmentation networks~\cite{SegNet, DeeplabV3, PSPNet, DANet, CCNet, CGNet, BiSeNetV2, SegFormer} are directly applied and fitted in well on RS images. 
\par Compared to natural scene images, remote sensing images contain more specific detailed information like element boundary element corner, irregular shape, etc. Moreover, large geographical coverage and confusing geographical elements on RS images cause large intra-class variance and inter-class similarity. On solving these problems, many novel and strong RS segmentation networks are proposed. \cite{RSSeg-AMDF} propose a novel adaptive multi-scale deep fusion (AMDF) ResNet to fuse multiple hierarchy features in an adaptive manner. With abundant and discriminative feature extraction, AMDF-ResNet achieves high-level performance. \cite{RSSeg-DMSeg} propose an efficient dilated network, which improves the network by exploring the multi-context features. Besides exploiting the effective information on multiple features, some works focus on integrating attention mechanisms into RS segmentation networks. \cite{RSSeg-Relation, RSSeg-SCAttNet, RSSeg-SSAtNet} utilize spatial-wise and channel-wise attention modules, which can overcome misunderstanding by capturing long-range spatial relationships and finding important channels. \cite{RSSeg-ERN, RSSeg-EANet, RSSeg-MSBANet, RSSeg-BSNet} all pay attention to boundary parsing. These edge-sensitive architectures offer an inspiring perspective on enhancing the understanding of complex geographical elements in pixel-wise.
\subsection{Unsupervised Cross-domain Adaptation for Semantic Segmentation}
\par With domain shift alleviating mechanism, unsupervised domain adaptation (UDA) techniques can make source-trained (trained only with source samples) models adapt to target samples. Recently, many remarkable works have made huge progress in applying UDA to cross-domain semantic segmentation tasks. 
\par Adversarial learning is frequently adopted in many excellent methods. Some methods utilize image generalization techniques like image translation~\cite{cyclegan, DA-pix2pix} to align image appearance between source and target images. \cite{DA-GIO, DA-FDA, DA-LFG, DA-SIFA, DA-DSFN} employ image-level adaption in the first step and then train the segmentation networks with cross-domain synthetic data. Other methods explore the domain-invariant features between source-style and target-style features, which takes advantage of the feature-level alignment strategy to solve the domain shift problem. \cite{DA-AdapSegNet, DA-SSF-DAN, DA-CM, DA-EGUDA} insert discriminators into networks for consistency alignment on intermediate feature maps or output entropy maps.
\par As another typical non-adversarial UDA paradigm, self-training has attracted much attention in cross-domain semantic segmentation tasks. \cite{DA-CRST, DA-DomainAS, UASR, DAFormer} promote the adaption ability by generating reliable, consistent, and class-balanced pseudo labels. Supervised by target pseudo ground-truth, models can quickly adapt to target images.
\subsection{Cross-domain Semantic Segmentation on RS Images}
\par Although the cross-domain RS image semantic segmentation task has not been fully studied and the cross-domain adaptive potential has not been fully exploited, there are still many awesome works proposed in recent years. \cite{DA-UDAGAN} first address the domain adaptation issue for the RS image semantic segmentation task. They use generative adversarial networks (GANs) based architecture to tackle this task and achieve convincing results. Following their pioneer work, \cite{DA-DualGAN} introduce DualGAN~\cite{DualGAN} and demonstrate its strong adaptation ability on the RS image semantic segmentation task. Similarly, \cite{DA-UDA1, DA-UDA2, DA-ResiDualGAN} all design GANs-based networks and explore the adaptive potential of GANs on cross-domain RS image semantic segmentation tasks. \cite{DA-CLAL, Maetal-MBATA, Wangetal-FGUDA} propose a novel network integrating adversarial learning and contrastive learning. Combined with adversarial loss and pixel-wise contrastive loss, the segmentation network can learn rich domain-invariant features. \cite{DA-DeepCA} also focus on extracting domain-invariant features. Instructed by this principle, they propose a deep covariance alignment (DCA) module, which achieves competitive results on some popular benchmarks, LoveDA~\cite{LoveDA}. \cite{DA-SDA} propose a step-wise RS segmentation network with covariate shift alleviation to close the gap between source and target domains. \cite{DA-CSLG} a local-to-global RS segmentation framework that follows a curriculum-style approach. They design a two-stage cross-domain adaptation: ``source domain to Easy-to-adapt'' and ``Easy-to-adapt to Hard-to-adapt''. All above-mentioned brilliant works advance the cross-domain RS image semantic segmentation task.
\begin{figure*}
\begin{center}
   \includegraphics[width=1.0\linewidth]{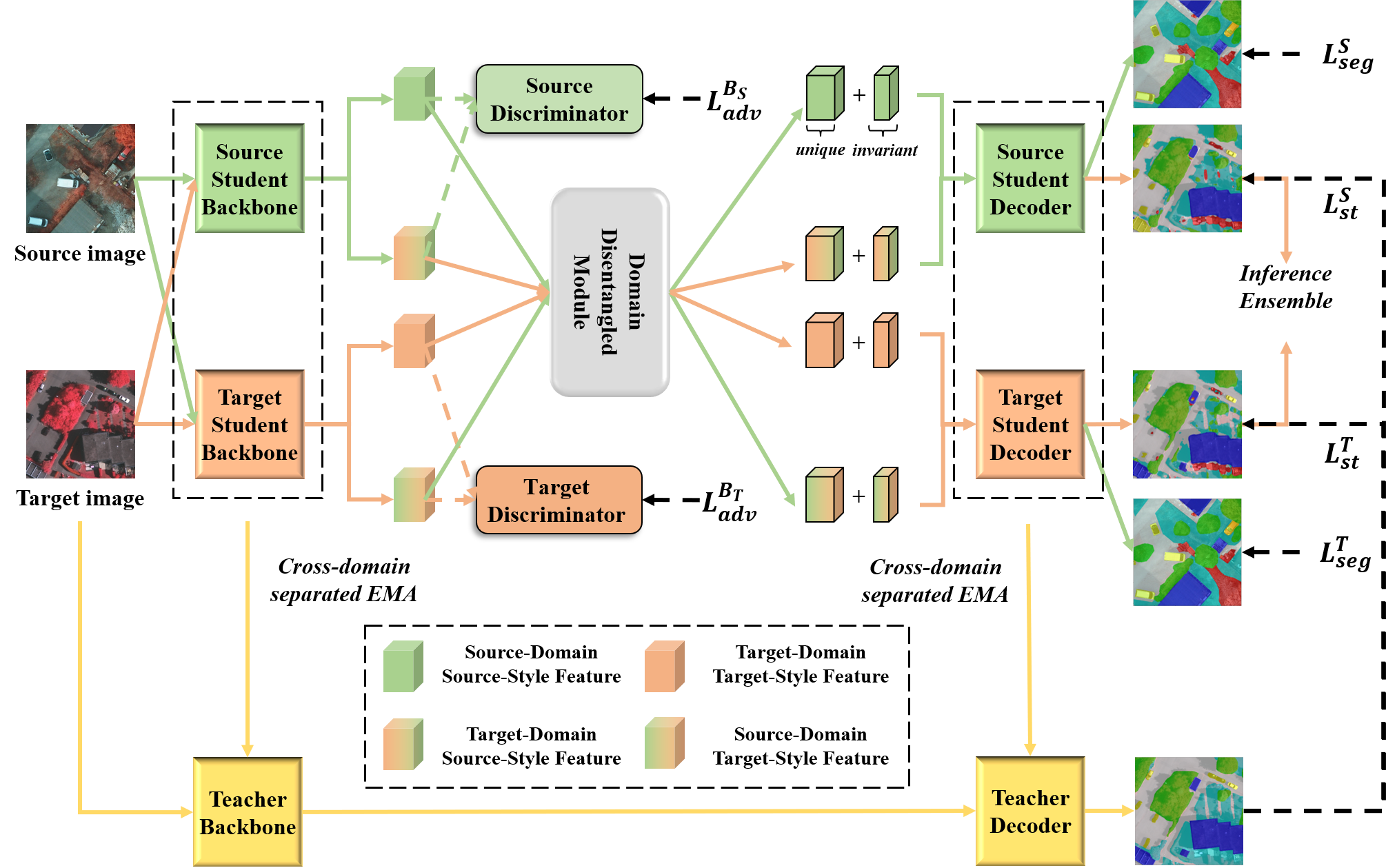}
\end{center}
   \caption{\textbf{The overview of ST-DASegNet.} In this architecture, source, and target student backbones are designed for feature extraction on two-style features. Domain disentangled module is designed to extract the cross-domain universal features and single-domain unique features. Source and target student decoders are designed to respectively map source and target features into predictions. Teacher backbone and decoder are respectively updated by student backbones and student decoders with the EMA technique. $D_{S}$ and $D_{T}$ are two discriminators to align cross-domain single-style features.}
\label{Fig2}
\end{figure*}
\section{Proposed Method} \label{sec:Method}
\par To decrease the negative influence of domain shift between source and target RS images, we propose ST-DASegNet. Fig.~\ref{Fig2} shows the overview of ST-DASegNet. In this paper, we follow the unsupervised domain adaptation constraint, which means annotations of source samples are given while no annotation of target samples is available.
\subsection{Source and Target Student Backbones}
\par From our proposed first principle in Sec.\ref{sec::intro}, we believe that unique features lead to large visual discrepancy while invariant features lead to easy identification. Naturally, if the model can precisely extract two-style features from two-domain images, it will benefit the unique and invariant feature representation. Motivated by this reasonable assumption, we design two student backbones rather than common-applied weight-sharing backbone. Here, the source student backbone ($B_{S}$) is expected to extract source-style features from two-domain images. Similarly, the target student backbone ($B_{T}$) is expected to extract target-style features from two-domain images.
\par As shown in Fig.~\ref{Fig2}, source and target images ($\bm{x_{s}}$ and $\bm{x_{t}}$) will be respectively fed into the source and target student backbones. Eq.~\ref{Eq1} and Eq.~\ref{Eq2} show this process.
\begin{equation}
\bm{F_{S-s}} = B_{S}(\bm{x_{s}}),\quad \bm{F_{S-t}} = B_{S}(\bm{x_{t}})
\label{Eq1}
\end{equation}
\begin{equation}
\bm{F_{T-s}} = B_{T}(\bm{x_{s}}),\quad \bm{F_{T-t}} = B_{T}(\bm{x_{t}})
\label{Eq2}
\end{equation}
\par where $\bm{F_{S-s}}$ and $\bm{F_{S-t}}$ are two output features from source student backbone, which respectively denote source-domain source-style feature and target-domain source-style feature. $\bm{F_{T-s}}$ and $\bm{F_{T-t}}$ are two output features from target student backbone, which respectively denote source-domain target-style feature and target-domain target-style feature. 
\subsection{Adversarial Learning on Cross-Domain Single-Style Features}
\par From our proposed second principle in Sec.\ref{sec::intro}, we believe that feature-level alignment may be more suitable than image-level alignment on cross-domain RS image semantic segmentation task. Feature alignment can guarantee the basic adaptation ability of student backbones and enhance the representation consistency between cross-domain images. In this paper, we design two discriminators to apply adversarial learning on cross-domain single-style features. Specifically, source discriminator ($D_{S}$) is proposed to align source-domain source-style feature and target-domain source-style feature. target discriminator ($D_{T}$) is proposed to align source-domain target-style feature and target-domain target-style feature. The mathematical operations are shown in Eq.~\ref{Eq3} and Eq.~\ref{Eq4}.
\begin{equation}
\begin{split}
   \mathcal{L}_{adv}^{B_{S}}(B_{S}, D_{S}) &= \mathbb{E}_{x^{s} \sim X^{s}}[log(D_{S}(\bm{F_{S-s}}))]  \\
                  &+ \mathbb{E}_{x^{t} \sim X^{t}}[log(1 - D_{S}(\bm{F_{S-t}}))]
   \end{split}
\label{Eq3}
\end{equation}
\begin{equation}
\begin{split}
   \mathcal{L}_{adv}^{B_{T}}(B_{T}, D_{T}) &= \mathbb{E}_{x^{t} \sim X^{t}}[log(D_{T}(\bm{F_{T-t}}))]  \\
                  &+ \mathbb{E}_{x^{s} \sim X^{s}}[log(1 - D_{T}(\bm{F_{T-s}}))]
   \end{split}
\label{Eq4}
\end{equation}
\par where $\mathcal{L}_{adv}^{B_{S}}(B_{S}, D_{S})$ and $\mathcal{L}_{adv}^{B_{T}}(B_{T}, D_{T})$ are adversarial losses. To optimize $B_{S}$ and $B_{T}$. We adopt ``min-max'' criterion, which can be expressed in Eq.~\ref{Eq5}.
\begin{equation}
\begin{cases}{}
    min_{B_{S}}max_{D_{S}}\mathcal{L}_{adv}^{B_{S}}(B_{S}, D_{S}) \\ \\ min_{B_{T}}max_{D_{T}}\mathcal{L}_{adv}^{B_{T}}(B_{T}, D_{T})
\end{cases}
\label{Eq5}
\end{equation}
\begin{figure*}
\begin{center}
   \includegraphics[width=1.0\linewidth]{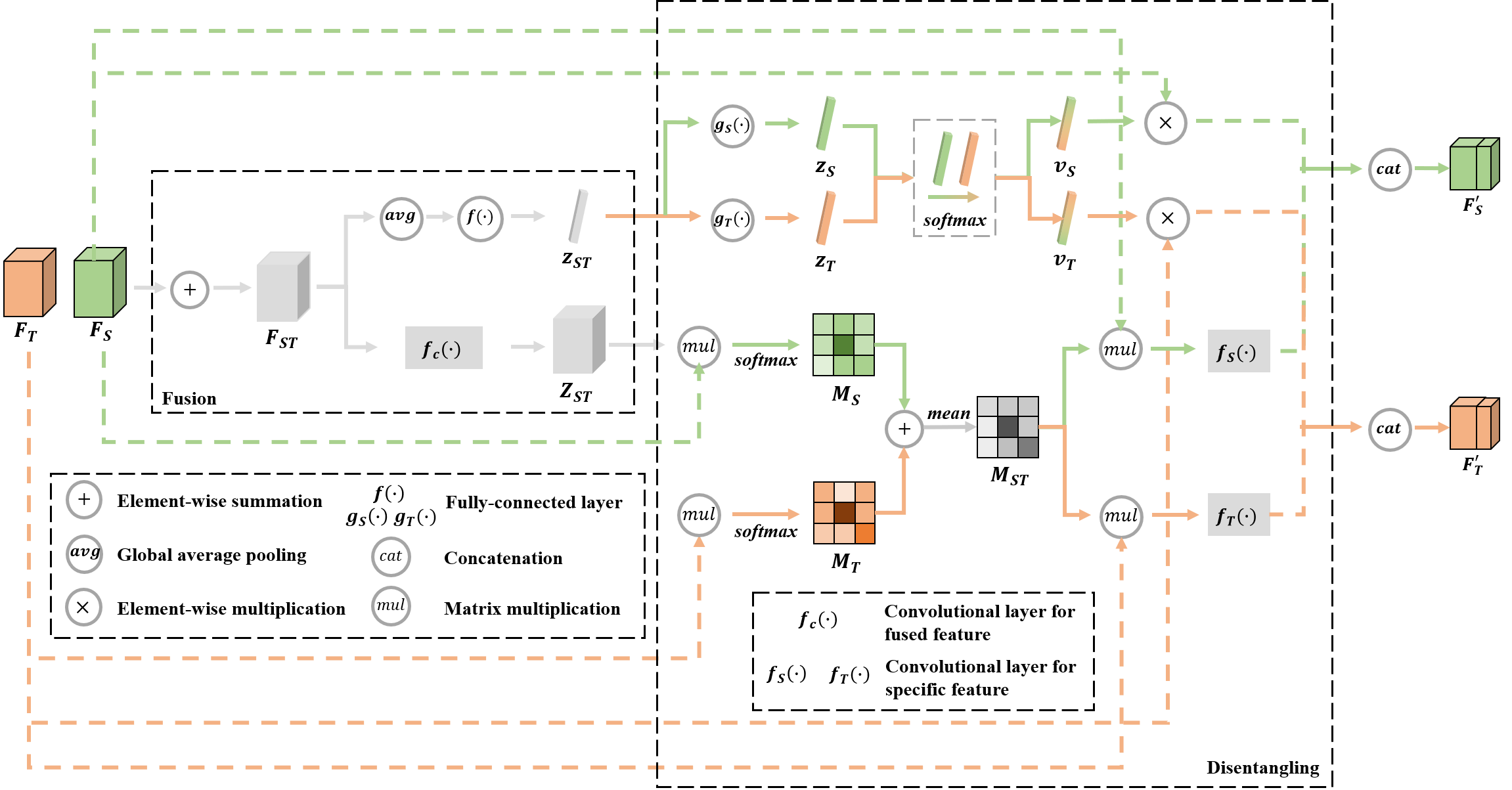}
\end{center}
   \caption{\textbf{The module structure of domain disentangled module.} In this module, $\bm{F_{S}}$ and $\bm{F_{T}}$ respectively denote source-style and target-style features. Particularly, two features are extracted from the single-domain image (e.g., $\bm{F_{S-s}}$ and $\bm{F_{T-s}}$). This module consists of fusion and disentangling blocks. A fusion block is used to generate a fused prototype and feature map. In the disentangling block, a fused prototype is used to generate unique features while a fused feature map is used to generate invariant features. Finally, these two features are concatenated together.}
\label{Fig3}
\end{figure*}
\subsection{Domain Disentangled Module on Single-Domain Cross-Style Features}
\par From source and target student backbones, we can obtain two-style features for source and target images. With feature-level adversarial learning on these features, the domain shift problem is alleviated. Naturally, we prepare to extract universal and unique features from single-domain cross-style features. In this paper, we propose a domain disentangled module (DDM). The module structure is shown in Fig.~\ref{Fig3}. 
\par In DDM, two input features are extracted by two student backbones from the single domain image, so two features have different styles. Here, we propose a fusion block to generate a fused prototype ($\bm{z_{ST}}$) and feature map ($\bm{Z_{ST}}$). The forward process of the fusion block is shown in Eq.~\ref{Eq6}. 
\begin{equation}
    \bm{z_{ST}} = f(avg(\bm{F_{ST}})), \quad \bm{Z_{ST}} = f_{c}(\bm{F_{ST}})
\label{Eq6}    
\end{equation}
\par As shown in Fig.~\ref{Fig3}, $f(\cdot)$ denotes a fully-connected layer (dimensionality-reduction). $f_{c}(\cdot)$ denotes a convolutional layer. $avg(\cdot)$ denotes channel-wise global average pooling operation. $\bm{F_{ST}} = \bm{F_{S}} + \bm{F_{T}}$.
\par We then propose a disentangling block to decouple unique and invariant features. This block can be separated into two parts. The first part is designed to extract unique features. Even though source and target student backbones can represent single-domain images as source and target styles, the output features are still mixtures partially containing other style features (E.g., source-domain source-style feature contains some target-style information). Therefore, unique feature extraction can be regarded as purifying the style-specific feature from the fused feature. As shown in Fig.\ref{Fig3}, $\bm{z_{ST}}$ is served as a guided prototype to generate two complementary vectors ($\bm{v_{S}}$ and $\bm{v_{T}}$). The generation process is shown in Eq.~\ref{Eq7} and Eq.~\ref{Eq8}.
\begin{equation}
    \bm{v_{S}} = \frac{exp(\bm{z_{S}})}{exp(\bm{z_{S}}) + exp(\bm{z_{T}})}, \ \, \bm{v_{T}} = \frac{exp(\bm{z_{T}})}{exp(\bm{z_{S}}) + exp(\bm{z_{T}})}
\label{Eq7}    
\end{equation}
\begin{equation}
    \bm{z_{S}} = g_{S}(\bm{z_{ST}}), \quad \bm{z_{T}} = g_{T}(\bm{z_{ST}}) 
\label{Eq8}    
\end{equation}
\par where $g_{S}(\cdot)$ and $g_{T}(\cdot)$ are two style-specific fully-connected layers (dimensionality-increase). $\bm{v_{S}} + \bm{v_{T}} = \bm{1}$. Eq.~\ref{Eq7} shows $softmax$ operation.
\par With $\bm{v_{S}}$ and $\bm{v_{T}}$, we conduct channel-wise multiplication on input features ($\bm{F_{S}}$ and $\bm{F_{T}}$). Here, $\bm{v_{S}} = [v_{S}^{1}, v_{S}^{2}, \cdots, v_{S}^{C}]$. $\bm{v_{T}} = [v_{T}^{1}, v_{T}^{2}, \cdots, v_{T}^{C}]$. $\bm{F_{S}} = [\bm{F_{S}^{1}}, \bm{F_{S}^{2}}, \cdots, \bm{F_{S}^{C}}$]. $\bm{F_{T}} = [\bm{F_{T}^{1}}, \bm{F_{T}^{2}}, \cdots, \bm{F_{T}^{C}}$]. Eq.~\ref{Eq9} shows the channel-wise multiplication operation.
\begin{equation}
    \bm{F_{U-S}^{i}} = \bm{F_{S}^{i}} \times v_{S}^{i}, \quad \bm{F_{U-T}^{i}} = \bm{F_{T}^{i}} \times v_{T}^{i}
\label{Eq9}    
\end{equation}
\par where $v_{S}^{i}$ and $v_{T}^{i}$ are scalars of $\bm{v_{S}}$ and $\bm{v_{T}}$ ($\mathbb{R}^{C}$). $\bm{F_{S}^{i}}$ and $\bm{F_{T}^{i}}$ ($\mathbb{R}^{H \times W}$) are channels of $\bm{F_{S}}$ and $\bm{F_{T}}$ ($\mathbb{R}^{C \times H \times W}$). $\bm{F_{U-S}}$ and $\bm{F_{U-T}}$ are respectively source-style and target-style unique features.  
\par The second part of DDM is designed to extract invariant features. The key idea of this part is to find the invariant channel-wise relation mask between input features ($\bm{F_{S}}$ and $\bm{F_{T}}$) and fused feature map ($\bm{Z_{ST}} = [\bm{Z_{ST}^{1}}, \bm{Z_{ST}^{2}}, \cdots, \bm{Z_{ST}^{C}}]$). Here, we reshape each element ($\bm{F_{S}^{i}}$ and $\bm{F_{T}^{i}}$) of $\bm{F_{S}}$ and $\bm{F_{T}}$ to $\mathbb{R}^{N}$, where $N = H \times W$. Similarly, we reshape $\bm{Z_{ST}^{i}}$ to $\mathbb{R}^{N}$. After reshaping, we apply dot product with $softmax$ operation, which is shown in Eq.~\ref{Eq10} $\sim$ Eq.~\ref{Eq12}.
\begin{equation}
    M_{S}^{j, i} = \frac{exp(\bm{F_{S}^{i}} \cdot \bm{Z_{ST}^{j}})}{\sum_{i=1}^{C}(\bm{F_{S}^{i}} \cdot \bm{Z_{ST}^{j}})}, \ \, M_{T}^{j, i} = \frac{exp(\bm{F_{T}^{i}} \cdot \bm{Z_{ST}^{j}})}{\sum_{i=1}^{C}(\bm{F_{T}^{i}} \cdot \bm{Z_{ST}^{j}})}
\label{Eq10}  
\end{equation}
\begin{equation}
    {\bm{M}_{S}=\begin{bmatrix}
    {M}_{S}^{1, 1} & \dots & {M}_{S}^{1, C}\\
     \vdots & \ddots & \vdots\\
    {M}_{S}^{C, 1} & \dots & {M}_{S}^{C, C} 
    \end{bmatrix}}
\label{Eq11}
\end{equation}
\begin{equation}
    {\bm{M}_{T}=\begin{bmatrix}
    {M}_{T}^{1, 1} & \dots & {M}_{T}^{1, C}\\
     \vdots & \ddots & \vdots\\
    {M}_{T}^{C, 1} & \dots & {M}_{T}^{C, C} 
    \end{bmatrix}}
\label{Eq12}
\end{equation}
\par where $\bm{M_{S}}$ ($\mathbb{R}^{C \times C}$) represents the channel-wise relation mask between $\bm{F_{S}}$ and $\bm{Z_{ST}}$ while $\bm{M_{T}}$ ($\mathbb{R}^{C \times C}$) represents the channel-wise relation mask between $\bm{F_{T}}$ and $\bm{Z_{ST}}$. The invariant relation mask is denoted as $\bm{M_{ST}} = (\bm{M_{S}} + \bm{M_{T}}) / 2$. In $\bm{M_{ST}}$, if the value ($\bm{M_{ST}^{j, i}}$) is larger, it means that the $i^{th}$ channel of $\bm{F_{S}}$ and $\bm{F_{T}}$ probably both have high impact on the $j^{th}$ channel of $\bm{M_{ST}}$.
\par Since $\bm{M_{ST}}$ reflects the universal relation, we perform a matrix multiplication between $\bm{M_{ST}}$ and $\bm{F_{S}}, \bm{F_{T}}$ to generate features with channel-wise attention. Eq.~\ref{Eq13} shows this process.
\begin{equation}
    \bm{F_{I-S}^{j}} = \sum_{i=1}^{C}(\bm{F_{S}^{i}} \times M_{ST}^{j, i}), \quad \bm{F_{I-T}^{j}} = \sum_{i=1}^{C}(\bm{F_{T}^{i}} \times M_{ST}^{j, i})
\label{Eq13}
\end{equation}
\ where $\bm{F_{I-S}} = [\bm{F_{I-S}^{1}}, \bm{F_{I-S}^{2}}, \cdots, \bm{F_{I-S}^{C}}]$ and $\bm{F_{I-T}} = [\bm{F_{I-T}^{1}}, \bm{F_{I-T}^{2}}, \cdots, \bm{F_{I-T}^{C}}]$ are respectively source-style and target-style invariant features.
\par Finally, we fuse the unique and invariant feature through concatenation. The outputs of DDM are $\bm{F_{S}^{'}}$ and $\bm{F_{T}^{'}}$. The pipeline of DDM is shown in Alg.\ref{Alg1}.
\begin{algorithm}[H]
\normalsize
		\caption{Domain disentangled module}
		\label{Alg1}
		\begin{algorithmic}
			\REQUIRE Source-style feature map $\bm{F_{S}}$ and target-style feature map $\bm{F_{T}}$ extracted from single-domain image. $\bm{F_{S}}, \bm{F_{T}} \in \mathbb{R}^{C \times H \times W}$. Trainable parameters in fully-connected layers ($f(\cdot)$, $g_{S}(\cdot)$, $g_{T}(\cdot)$) and convolutional layers ($f_{c}(\cdot)$, $f_{S}(\cdot)$, $f_{T}(\cdot)$)
			\ENSURE {New source-style feature map $\bm{F_{S}^{'}}$ and target-style feature map $\bm{F_{T}^{'}}$ containing unique and invariant information. $\bm{F_{S}^{'}}, \bm{F_{T}^{'}} \in \mathbb{R}^{C \times H \times W}$}.
			\STATE {\textbf{Feature Fusion:}}
			\STATE {Get fused prototype $\bm{z_{ST}}$ using Eq.~\ref{Eq6}.}
			\STATE {Get fused feature map $\bm{Z_{ST}}$ using Eq.~\ref{Eq6}.}
			\STATE {\textbf{Feature Disentangling:}}
			\INDSTATE {\textbf{Unique feature disentangling:}}
			\INDSTATE {Get complementary channel-wise weighted vectors $\bm{v_S}$ and $\bm{v_T}$ using Eq.~\ref{Eq7} and Eq.~\ref{Eq8}. }
			\INDSTATE {Get source-style and target-style unique features $\bm{F_{U-S}}$ and $\bm{F_{U-T}}$ using Eq.~\ref{Eq9}.}
			\INDSTATE {\textbf{Invariant feature disentangling:}}
			\INDSTATE {Get channel-wise relation masks $\bm{M_{S}}$ and $\bm{M_{T}}$ using Eq.~\ref{Eq10} $\sim$ Eq.~\ref{Eq12}.}
			\INDSTATE {Get source-style and target-style invariant features $\bm{F_{I-S}}$ and $\bm{F_{I-T}}$ using Eq.~\ref{Eq13}.}
			\INDSTATE {\textbf{Feature concatenation:}}
			\INDSTATE {Get source-style new features $\bm{F_{S}^{'}} = cat(\bm{F_{U-S}}, \bm{F_{I-S}})$.}
			\INDSTATE {Get target-style new features $\bm{F_{T}^{'}} = cat(\bm{F_{U-T}}, \bm{F_{I-T}})$.}
		\end{algorithmic}
\end{algorithm}
\begin{figure*}
\begin{center}
   \includegraphics[width=1.0\linewidth]{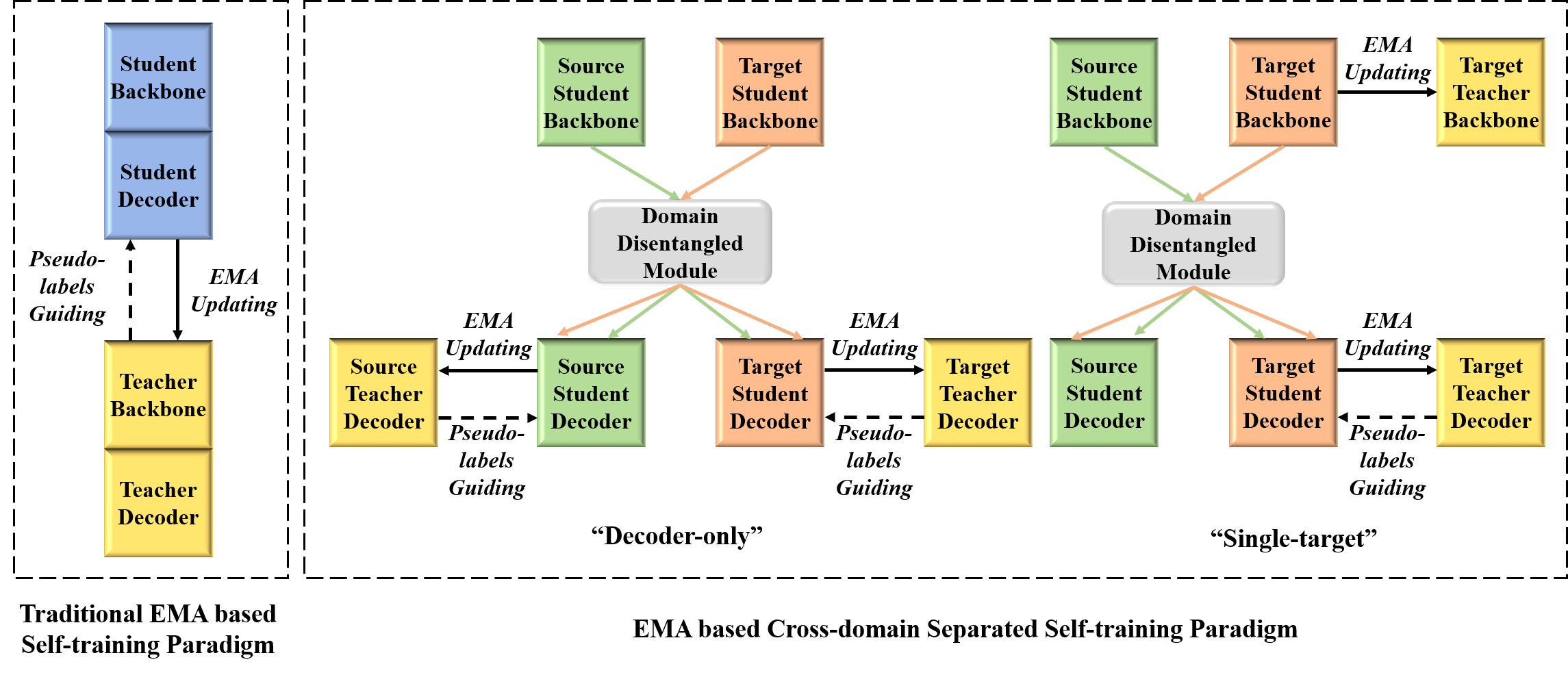}
\end{center}
   \caption{Comparison between traditional EMA-based self-training paradigm and our proposed EMA-based cross-domain separated self-training paradigm.}
\label{Fig4}
\end{figure*}
\subsection{EMA-based Cross-Domain Separated Self-Training Mechanism}
\par From our proposed third and fourth principles in Sec.\ref{sec::intro}, we believe that directly applying an adversarial learning mechanism can ease the domain shift, but it will still cause instability and representation tendency on source images because of no target annotations. Motivated by further improving the representation capability on target images, we propose an EMA-based cross-domain separated self-training mechanism. As shown in Fig.~\ref{Fig4}, the traditional EMA-based self-training paradigm generates a teacher network with EMA updating operation on the student network. Compared to traditional paradigm, our self-training paradigm will not generate a whole teacher network, but only generate some components. In this paper, we propose two self-training paradigms, which are ``Decoder-only'' and ``Single-target''.
\subsubsection{Decoder-only} As shown in Fig.~\ref{Fig4}, ``Decoder-only'' paradigm introduces two teacher decoders rather than two teacher networks. In this paradigm, student networks get credible and mature knowledge from teacher decoders. Specifically, we first construct two teacher decoders, which are the source teacher decoder ($H_{S}^{te}$) and the target teacher decoder ($H_{T}^{te}$). Then, at training step $t$, the teacher decoders' weights are updated by student decoders' weights by EMA operation, which can be formulated in Eq.~\ref{Eq14}.
\begin{equation}
\begin{cases}{}
    \phi_{t}^{H_{S}^{te}} = \alpha\phi_{t-1}^{H_{S}^{te}} + (1-\alpha)\theta_{t}^{H_{S}} \\ \\ \phi_{t}^{H_{T}^{te}} = \alpha\phi_{t-1}^{H_{T}^{te}} + (1-\alpha)\theta_{t}^{H_{T}}
\end{cases}
\label{Eq14}
\end{equation}
\par where $\alpha$ denotes the EMA decay factors controlling the updating rate. $\phi_{t}^{H_{S}^{te}}, \phi_{t}^{H_{T}^{te}}$ refer to weights of two teacher decoders at $t^{th}$ step. $\theta_{t}^{H_{S}}, \theta_{t}^{H_{T}}$ refer to weights of two student decoders ($H_{S}$, $H_{T}$) at $t^{th}$ step. 
\par After updating the weights of teacher decoders, we transfer the reliable knowledge to student by pseudo-label guidance. With two teacher decoders, the generation of pseudo-label is formulated in Eq.~\ref{Eq15}.
\begin{equation}
\bm{\hat{P}_{t}^{te}} = \mathop{\arg\max}\limits_{c} \frac{(H_{S}^{te}(\bm{F_{S-t}^{'}}) + H_{T}^{te}(\bm{F_{T-t}^{'}}))}{2}
\label{Eq15}
\end{equation}
\par where $\bm{F_{S-t}^{'}} \in \mathbb{R}^{C \times H \times W}$ and $\bm{F_{T-t}^{'}} \in \mathbb{R}^{C \times H \times W}$ are respectively target-domain source-style feature and target-domain target-style feature after DDM (Alg.\ref{Alg1}). $\bm{\hat{P}_{t}^{te}} \in \mathbb{R}^{H \times W}$ is pseudo-label (index map) after channel-wise $argmax$ operation. Here, we apply a soft-voting ensemble strategy to integrate the predictions after two teacher decoders, which makes the pseudo-label more credible.
\par ``Decoder-only'' paradigm has two main advantages. (1) It can efficiently transfer reliable knowledge to both source and target students (including backbone and decoder) for target-style feature extraction, which maximally alleviates the representation tendency on source images. (2) It can efficiently save a lot of training memory and computation cost because the decoder of a segmentation network is always much smaller than the backbone.
\subsubsection{Single-target} In this paper, we propose ``Single-target'' paradigm, which is another self-training paradigm on our proposed DASegNet. This paradigm mainly aims to enhance the target-style feature representation of the target student backbone and decoder. As shown in Fig.~\ref{Fig4}, we construct a whole target teacher network including target teacher backbone ($B_{T}^{te}$) and target teacher decoder ($H_{T}^{te}$). Similar to Eq.~\ref{Eq14}, we apply EMA updating on the target teacher network (Eq.~\ref{Eq16}). 
\begin{equation}
\begin{cases}{}
    \phi_{t}^{B_{T}^{te}} = \alpha\phi_{t-1}^{B_{T}^{te}} + (1-\alpha)\theta_{t}^{B_{T}} \\ \\ \phi_{t}^{H_{T}^{te}} = \alpha\phi_{t-1}^{H_{T}^{te}} + (1-\alpha)\theta_{t}^{H_{T}}
\end{cases}
\label{Eq16}
\end{equation}
\par where $\phi_{t}^{B_{T}^{te}}$ refers to weights of target teacher backbone at $t^{th}$ step. $\theta_{t}^{B_{T}}$ refers to weights of target student backbone at $t^{th}$ step. 
\par With a whole teacher network, the pseudo-label generation of ``Single-target'' paradigm can be expressed in Eq.~\ref{Eq17}.
\begin{equation}
\bm{\hat{P}_{t}^{te}} = \mathop{\arg\max}\limits_{c} H_{T}^{te}(\bm{F_{T-t}^{'}})
\label{Eq17}
\end{equation}
\par Compared to ``Decoder-only'' paradigm, ``Single-target'' paradigm only employs pseudo-label guiding on target students. Theoretically, ``Single-target'' paradigm can enhance the target-style feature extraction power on the target student, which leads to a larger representation gap between the source and student networks. In practical experiments, we find that ``Decoder-only'' paradigm can also enhance the target-style feature extraction power on target students with less memory and training computation cost. Basically, both these two paradigms improve the target-domain adaptation ability of DASegNet by credible pseudo-label guiding.
\par In summary, we show our proposed cross-domain separated self-training mechanism in Alg.\ref{Alg2}. In this paper, we mainly adopt ``Decoder-only'' paradigm. ``Single-target'' paradigm provides another perspective on integrating the self-training mechanism into DASegNet. In experiments, we will make further comparisons between these two paradigms.
\subsection{Optimization with Combined Loss}
\par In this paper, we apply segmentation loss, adversarial loss, and self-training loss to jointly optimize the ST-DASegNet. 
\par As for segmentation loss ($\mathcal{L}_{seg}^{S}$ and $\mathcal{L}_{seg}^{T}$) shown in Fig.~\ref{Fig2}, we apply conventional cross-entropy loss using annotations of source images. We formulate the segmentation loss function in Eq.~\ref{Eq18} and Eq.~\ref{Eq19}.
\begin{equation}
\begin{split}
    \mathcal{L}_{seg}^{S} = -\sum_{h=1}^{H}\sum_{w=1}^{W}\sum_{c=1}^{C}(\bm{y_{s}(h, w, c)}log(H_{S}(\bm{F_{S-s}^{'}}))) 
\end{split}
    \label{Eq18}
\end{equation}
\begin{equation}
\begin{split}
    \mathcal{L}_{seg}^{T} = -\sum_{h=1}^{H}\sum_{w=1}^{W}\sum_{c=1}^{C}(\bm{y_{s}(h, w, c)}log(H_{T}(\bm{F_{T-s}^{'}})))
\end{split}
    \label{Eq19}
\end{equation}
\par where, $\bm{y_{s}}$ is the ground truth of source images. $\bm{F_{S-s}^{'}}$ and $\bm{F_{T-s}^{'}}$ are respectively source-domain source-style and source-domain target-style disentangled features after DDM.
\par As for self-training loss ($\mathcal{L}_{st}^{S}$ and $\mathcal{L}_{st}^{T}$) shown in Fig.~\ref{Fig2}, we also adopt cross-entropy loss using pseudo-labels of target images. We formulate self-training loss function in Eq.~\ref{Eq20} and Eq.~\ref{Eq21}.
\begin{equation}
\begin{split}
    \mathcal{L}_{st}^{S} = -\sum_{h=1}^{H}\sum_{w=1}^{W}\sum_{c=1}^{C}(\bm{\hat{P}_{t}^{te}(h, w, c)}log(H_{S}(\bm{F_{S-t}^{'}}))) 
\end{split}
    \label{Eq20}
\end{equation}
\begin{equation}
\begin{split}
    \mathcal{L}_{st}^{T} = -\sum_{h=1}^{H}\sum_{w=1}^{W}\sum_{c=1}^{C}(\bm{\hat{P}_{t}^{te}(h, w, c)}log(H_{T}(\bm{F_{T-t}^{'}})))
\end{split}
    \label{Eq21}
\end{equation}
\par where, $\bm{\hat{P}_{t}^{te}}$ denotes the pseudo-labels of target images, which are transformed from index maps ($\mathbb{R}^{H \times W}$) to one-hot labels ($\mathbb{R}^{C \times H \times W}$). The calculation operation of $\bm{\hat{P}_{t}^{te}}$ is shown in Eq.~\ref{Eq15} or Eq.~\ref{Eq17} respectively for ``Decoder-only'' and ``Single-target'' paradigms.
\par To optimize ST-DASegNet, we combine three types of losses together. The final combined loss function ($\mathcal{L}$) is formulated in Eq.~\ref{Eq22}.
\begin{equation}
    \mathcal{L} =  \mathcal{L}_{seg} + \lambda \mathcal{L}_{st} + \beta \mathcal{L}_{adv}
    \label{Eq22}
\end{equation}
\par where $\mathcal{L}_{seg} = \mathcal{L}_{seg}^{S} + \mathcal{L}_{seg}^{T}$ (Eq.~\ref{Eq18} and Eq.~\ref{Eq19}). $\mathcal{L}_{st} = \mathcal{L}_{st}^{S} + \mathcal{L}_{st}^{T}$ (Eq.~\ref{Eq20} and Eq.~\ref{Eq21}). $\mathcal{L}_{adv} = \mathcal{L}_{adv}^{B_{S}} + \mathcal{L}_{adv}^{B_{T}}$ (Eq.~\ref{Eq3} and Eq.~\ref{Eq4}). $\lambda$ and $\beta$ are two factors to adjust the proportion of self-training loss and adversarial loss. In this paper, we set $\lambda = 0.25$ and $\beta = 0.005$.
\begin{algorithm}[H]
\normalsize
		\caption{Cross-domain separated self-training mechanism}
		\label{Alg2}
		\begin{algorithmic}
			\REQUIRE Target images $\bm{x_{t}}$. Source student backbone $B_{S}$, target student backbone $B_{T}$, source student decoder $H_{S}$ and target student decoder $H_{T}$.
			\ENSURE {Trainable parameters of teacher network components at $t^{th}$ step. Pseudo-label $\bm{\hat{P}_{t}^{te}}$}.
			\IF {Self-training paradigm is ``Decoder-only''}
			\STATE {\textbf{EMA Updataing:}}
			\STATE {Update source teacher decoder $H_{S}^{te}$ and target teacher decoder $H_{T}^{te}$ at $t^{th}$ step using Eq.~\ref{Eq14}.}
			\STATE {\textbf{Pseudo-label Generation:}}
			\STATE {Extract features ($\bm{F_{S-t}}$ , $\bm{F_{T-t}}$) on $\bm{x_{t}}$ respectively from $B_{S}$ and $B_{T}$ using Eq.~\ref{Eq1} and Eq.\ref{Eq2}}
			\STATE {Get disentangle features ($\bm{F_{S-t}^{'}}$ , $\bm{F_{T-t}^{'}}$) from DDM with $\bm{F_{S-t}}$ and $\bm{F_{T-t}}$ as input using Alg.\ref{Alg1}.}
			\STATE {Get pseudo-label $\bm{\hat{P}_{t}^{te}}$ with $\bm{F_{S-t}^{'}}$ and $\bm{F_{T-t}^{'}}$ as input using Eq.~\ref{Eq15}.}
			\ENDIF
			\IF {Self-training paradigm is ``Single-target''}
			\STATE {\textbf{EMA Updataing:}}
			\STATE {Update target teacher backbone $B_{T}^{te}$ and target teacher decoder $H_{T}^{te}$ at $t^{th}$ step using Eq.~\ref{Eq16}.}
			\STATE {\textbf{Pseudo-label Generation:}}
			\STATE {Extract features ($\bm{F_{S-t}}$ , $\bm{F_{T-t}^{te}}$) on $\bm{x_{t}}$ respectively from $B_{S}$ and $B_{T}^{te}$ using Eq.~\ref{Eq1} and Eq.~\ref{Eq2}.}
			\STATE {Get disentangle features ($\bm{F_{S-t}^{'}}$ , $\bm{F_{T-t}^{'}}$) from DDM with $\bm{F_{S-t}}$ and $\bm{F_{T-t}^{te}}$ as input using Alg.\ref{Alg1}.}
			\STATE {Get pseudo-label $\bm{\hat{P}_{t}^{te}}$ with $\bm{F_{T-t}^{'}}$ as input using Eq.~\ref{Eq17}.}
			\ENDIF
		\end{algorithmic}
\end{algorithm}
\section{Experiments and Analysis}
\subsection{Datasets and Evaluation Metric}
\par To evaluate our method on cross-domain RS image semantic segmentation task, we use three benchmark datasets, which are Potsdam, Vaihingen, and LoveDA.
\par \textbf{Potsdam and Vaihingen.} These two datasets belong to ISPRS 2D semantic segmentation benchmark dataset~\cite{ISPRS}. The Potsdam dataset contains 38 VHR TOP (very-high-resolution True Orthophotos) with the size of 6000 $\times$ 6000 (fixed size). The Potsdam dataset has three different imaging modes, which are IR-R-G, R-G-B, and R-G-B-IR. The first two modes are 3-channel while the last mode is 4-channel. In this paper, we choose to use the first two modes. The Vaihingen dataset contains 33 VHR TOP with the size of 2000 $\times$ 2000 (approximate size). The Vaihingen dataset only has one imaging mode, which is IR-R-G. For more efficient computation cost, we crop the images of these two datasets into smaller patches with the size of 512 $\times$ 512. Specifically, we respectively select 512 and 256 as cropping strides for Potsdam and Vaihingen, generating 4598 and 1696 patches. Moreover, we split the Potsdam and Vaihingen datasets into training and testing sets. For Potsdam, the training and testing set contains 2904 and 1694 images, respectively. For Vaihingen, the training and testing set contain 1296 and 440 images, respectively. It is worth noting that, all the data preprocessing methods followed previous works~\cite{DA-DualGAN, DA-CLAL, DA-UDA1, DA-CSLG, DA-ResiDualGAN}.
\par On Potsdam/Vaihingen datasets, we design four cross-domain RS semantic segmentation tasks, which are listed as follows.
\begin{itemize}
\item Potsdam IR-R-G to Vaihingen IR-R-G (Potsdam IR-R-G $\rightarrow$ Vaihingen IR-R-G).
\item Vaihingen IR-R-G to Potsdam IR-R-G (Vaihingen IR-R-G $\rightarrow$ Potsdam IR-R-G).
\item Potsdam R-G-B to Vaihingen IR-R-G (Potsdam R-G-B $\rightarrow$ Vaihingen IR-R-G).
\item Vaihingen IR-R-G to Potsdam R-G-B (Vaihingen IR-R-G $\rightarrow$ Potsdam R-G-B).
\end{itemize}
\par \textbf{LoveDA.} This dataset is recently proposed to advance both RS semantic segmentation and domain adaptation tasks. It consists of 5987 high spatial resolution (1024 $\times$ 1024) RS images from three cities including Nanjing, Changzhou, and Wuhan. LoveDA dataset contains images from two domains (urban and rural), which focuses on challenging the model's generalized representation capacity on different geographical elements of urban and rural scenes. This dataset has 1833 urban images, which are split into 1156 training images and 677 validation images. For rural images, there are 2358 images in total, where 1366 images are used for training and the rest 992 images are used for validation. In addition, LoveDA also contains 1796 testing images (976 for rural and 820 for urban), which can be evaluated on online server\footnote{https://github.com/Junjue-Wang/LoveDA}. On LoveDA dataset, we design two cross-domain RS semantic segmentation tasks, which are urban-to-rural (urban $\rightarrow$ rural) and rural-to-urban (rural $\rightarrow$ urban) tasks. All our experiments setting is followed ~\cite{LoveDA, DA-DeepCA}.
\par \textbf{Evaluation Metric.} In this paper, we select common-used $IoU$ (intersection of union) and $F1$-score as evaluation metrics. Specifically, for a specific class $i$, $IoU$ is formulated as $IoU_{i} = tp_{i} / (tp_{i} + fp_{i} + fn_{i})$, where $tp_{i}$, $fp_{i}$, $fn_{i}$ denote true positive, false positive and false negative, respectively. The mIoU is the mean value of all categories' $IoU_{i}$. Additionally, $F1$-score is defined as $F1$-score $= (2 \times Precision \times Recall) / (Precision + Recall)$.
\subsection{Implementation Details}
\par Followed previous works, we select DeeplabV3~\cite{DeeplabV3} as baseline segmentation network. We also select a recent outstanding segmentation network, SegFormer~\cite{SegFormer} as another baseline. If applying DeeplabV3, the student backbone will be ResNet-50~\cite{ResNet} and the student decoder will be ASPP block (a combination of ASPP module and several convolutional blocks). To optimize DeeplabV3 based ST-DASegNet, we use SGD (Stochastic Gradient Descent) as an optimizer, where the initial learning rate is 0.001, the momentum value is 0.9 and the weight decay value is 0.0001. If applying SegFormer, the student backbone will be mit-b5~\cite{SegFormer} and the student decoder will be ``All MLP'' module. To optimize SegFormer based ST-DASegNet, we use Adam as optimizer and the initial learning rate is set as 0.0001. 
\par For DeeplabV3 and SegFormer based ST-DASegNet, we both apply Adam as an optimizer in which the initial learning rate is 0.00025. The structure of our proposed discriminators ($D_{S}$ and $D_{T}$) is followed PatchGAN~\cite{PatchGAN}. Specifically, the discriminator consists of 4 convolutional blocks with kernels as size of 4 $\times$ 4. The stride of the first two and the last two blocks is respectively set as 2 and 1. The output channels of each block are 64, 128, 256, and 1.
\par All experiments are implemented on mmsegmentation\footnote{https://github.com/open-mmlab/mmsegmentation} semantic segmentation framework and all models are trained on two NVIDIA RTX 3090. Our code is available at \url{https://github.com/cv516Buaa/ST-DASegNet}.
\begin{table*}	
	\centering
	\caption{Cross-domain RS image semantic segmentation comparison results (\%) from Potsdam IR-R-G to Vaihingen IR-R-G. Methods with ``*'' are our reimplemented version.}
    \scalebox{0.7}{
    \begin{tabular}{ccccccccccccccc}
		\cmidrule(r){1-15}
  \multirow{2}{*}{Methods} &  \multicolumn{2}{c}{Clutter} & \multicolumn{2}{c}{Impervious surfaces} & \multicolumn{2}{c}{Car} & \multicolumn{2}{c}{Tree} & \multicolumn{2}{c}{Low vegetation} & \multicolumn{2}{c}{Building} & \multicolumn{2}{c}{Overall}
  \\ \cmidrule(r){2-15}
  {} & $IoU$ & $F1$-score & $IoU$ & $F1$-score & $IoU$ & $F1$-score & $IoU$ & $F1$-score & $IoU$ & $F1$-score & $IoU$ & $F1$-score & $mIoU$ & $mF$-score 
  \\ \cmidrule(r){1-15}
  DeeplabV3~\cite{DeeplabV3} (Baseline) & 2.33 & 4.56 & 43.90 & 60.78 & 24.26 & 39.07 & 52.25 & 64.19 & 25.76 & 40.87 & 60.35 & 71.81 & 34.81 & 46.88 
  \\
  SegFormer~\cite{SegFormer} (Baseline) & 4.22 & 9.47 & 61.03 & 76.07 & 31.13 & 47.89 & 66.31 & 78.87 & 44.47 & 60.38 & 75.50 & 87.95 & 46.11 & 60.11
  \\
  \cmidrule(r){1-15}
  AdaptSegNet~\cite{AdapSegNet} & 4.60 & 8.76 & 54.39 & 70.39 & 6.40 & 11.99 & 52.65 & 68.96 & 28.98 & 44.91 & 63.14 & 77.40 & 35.02 & 47.05 
  \\
  FSDAN~\cite{FSDAN} & 10.00 & - & 57.40 & - & 37.00 & - & 58.40 & - & 41.70 & - & 57.80 & - & 43.70 & -
  \\
  ProDA~\cite{ProDA} & 3.99 & 8.21 & 62.51 & 76.85 & 39.20 & 56.52 & 56.26 & 72.09 & 34.49 & 51.65 & 71.61 & 82.95 & 44.68 & 58.05
  \\
  DualGAN~\cite{DA-DualGAN} & 29.66 & 45.65 & 49.41 & 66.13 & 34.34 & 51.09 & 57.66 & 73.14 & 38.87 & 55.97 & 62.30 & 76.77 & 45.38 & 61.43 
  \\
  Bai~\textit{et al.}~\cite{DA-CLAL} & 19.60 & 32.80 & 65.00 & 78.80 & 39.60 & 56.70 & 54.80 & 70.80 & 36.20 & 53.20 & 76.00 & 86.40 & 48.50 & 63.10
  \\
  Zhang~\textit{et al.}~\cite{DA-CSLG} & 20.71 & 31.34 & 67.74 & 80.13 & 44.90 & 61.94 & 55.03 & 71.90 & 47.02 & 64.16 & 76.75 & 86.65 & 52.03 & 66.02
  \\
  Wang~\textit{et al.}~\cite{DA-FGUDA} & 21.85 & 35.87 & \textbf{76.58} & \textbf{86.73} & 35.44 & 52.33 & 55.22 & 71.15 & 49.97 & 66.64 & 82.74 & 90.56 & 53.63 & 67.21 
  \\
  CIA-UDA~\cite{DA-CIAUDA} & 27.80 & 43.51 & 63.28 & 77.51 & 52.91 & 69.21 & 64.11 & 78.13 & 48.03 & 64.90 & 75.13 & 85.80 & 55.21 & 69.84
  \\ 
  ResiDualGAN~\cite{DA-ResiDualGAN} & 11.64 & 18.42 & 72.29 & 83.89 & \textbf{57.01} & \textbf{72.51} & 63.81 & 77.88 & 49.69 & 66.29 & 80.57 & 89.23 & 55.83 & 68.04
  \\ 
  DAFormer~\cite{DAFormer}* & 48.26 & 60.17 & 74.09 & 84.12 & 38.96 & 56.41 & \textbf{70.88} & \textbf{81.36} & \textbf{57.53} & \textbf{71.48} & 84.07 & 90.75 & 62.30 & 74.05
  \\ \cmidrule(r){1-15}
  ST-DASegNet (DeeplabV3) & 21.17 & 32.64 & 70.88 & 82.20 & 51.81 & 67.63 & 68.01 & 80.10 & 41.97 & 57.97 & 82.57 & 89.24 & 56.07 & 68.30
  \\
  ST-DASegNet (SegFormer) & \textbf{67.03} & \textbf{80.28} & 74.43 & 85.36 & 43.38 & 60.49 & 67.36 & 80.49 & 48.57 & 65.37 & \textbf{85.23} & \textbf{92.03} & \textbf{64.33} & \textbf{77.34}
  \\ \cmidrule(r){1-15}
   \end{tabular}
 }
 \label{Tab1}
 \end{table*}
 \begin{table*}	
	\centering
	\caption{Cross-domain RS image semantic segmentation comparison results (\%) from Vaihingen IR-R-G to Potsdam IR-R-G. Methods with ``*'' are our reimplemented version.}
    \scalebox{0.7}{
    \begin{tabular}{ccccccccccccccc}
		\cmidrule(r){1-15}
  \multirow{2}{*}{Methods} &  \multicolumn{2}{c}{Clutter} & \multicolumn{2}{c}{Impervious surfaces} & \multicolumn{2}{c}{Car} & \multicolumn{2}{c}{Tree} & \multicolumn{2}{c}{Low vegetation} & \multicolumn{2}{c}{Building} & \multicolumn{2}{c}{Overall}
  \\ \cmidrule(r){2-15}
  {} & $IoU$ & $F1$-score & $IoU$ & $F1$-score & $IoU$ & $F1$-score & $IoU$ & $F1$-score & $IoU$ & $F1$-score & $IoU$ & $F1$-score & $mIoU$ & $mF$-score 
  \\ \cmidrule(r){1-15}
  DeeplabV3~\cite{DeeplabV3} (Baseline) & 5.28 & 11.58 & 55.57 & 71.07 & 50.02 & 66.97 & 14.0 & 27.69 & 43.67 & 60.70 & 60.62 & 75.02 & 38.19 & 52.17
  \\
  SegFormer~\cite{SegFormer} (Baseline) & 1.08 & 2.65 & 60.63 & 76.47 & 58.99 & 73.14 & 30.07 & 46.24 & 51.91 & 68.82 & 74.85 & 87.18 & 46.29 & 59.08
  \\
  \cmidrule(r){1-15}
  AdaptSegNet~\cite{AdapSegNet} & 8.36 & 15.33 & 49.55 & 64.64 & 40.95 & 58.11 & 22.59 & 36.79 & 34.43 & 61.50 & 48.01 & 63.41 & 33.98 & 49.96
  \\
  ProDA~\cite{ProDA} & 10.63 & 19.21 & 44.70 & 61.72 & 46.78 & 63.74 & 31.59 & 48.02 & 40.55 & 57.71 & 56.85 & 72.49 & 38.51 & 53.82
  \\
  DualGAN~\cite{DA-DualGAN} & 11.48 & 20.56 & 51.01 & 67.53 & 48.49 & 65.31 & 34.98 & 51.82 & 36.50 & 53.48 & 53.37 & 69.59 & 39.30 & 54.71
  \\
  Zhang~\textit{et al.}~\cite{DA-CSLG} & \textbf{12.31} & \textbf{24.59} & 64.39 & 78.59 & 59.35 & 75.08 & 37.55 & 54.60 & 47.17 & 63.27 & 66.44 & 79.84 & 47.87 & 62.66 
  \\
  Wang~\textit{et al.}~\cite{DA-FGUDA} & 11.65 & 19.47 & 73.43 & 84.55 & 63.86 & 77.85 & 32.68 & 47.36 & 47.69 & 63.45 & 76.32 & 87.43 & 50.94 & 63.31 
  \\
  CIA-UDA~\cite{DA-CIAUDA} & 10.87 & 19.61 & 62.74 & 77.11 & 65.35 & 79.04 & 47.74 & 64.63 & 54.40 & 70.47 & 72.31 & 83.93 & 52.23 & 65.80
  \\ 
  DAFormer~\cite{DAFormer}* & 2.56 & 5.02 & 68.42 & 79.07 & 65.20 & 79.31 & \textbf{70.65} & \textbf{82.13} & 56.39 & 72.48 & 78.94 & 87.64 & 57.03 & 67.61
  \\ \cmidrule(r){1-15}
  ST-DASegNet (DeeplabV3) & 5.21 & 10.21 & 74.19 & 85.26 & \textbf{76.76} & \textbf{86.90} & 43.33 & 60.44 & 51.56 & 68.62 & 82.15 & 90.28 & 55.53 & 66.95
  \\
  ST-DASegNet (SegFormer) & 0.18 & 0.35 & \textbf{76.45} & \textbf{86.65} & 73.54 & 84.76 & 62.89 & 77.22 & \textbf{61.04} & \textbf{75.80} & \textbf{83.81} & \textbf{91.19} & \textbf{59.65} & \textbf{69.33}
  \\ \cmidrule(r){1-15}
   \end{tabular}
 }
 \label{Tab2}
 \end{table*}
 \subsection{Experimental Results}
\subsubsection{Cross-domain RS image semantic segmentation on Potsdam and Vaihingen} As mentioned above, we design four cross-domain tasks between Potsdam and Vaihingen. On these four tasks, we conduct abundant experiments to show the effectiveness of our proposed ST-DASegNet. Previous methods always select DeeplabV3 as the baseline model and hardly apply the transformer as the baseline model. Therefore, we reimplement DAFormer~\cite{DAFormer} to fairly compare with our SegFormer based ST-DASegNet. It is worth noting that DAFormer~\cite{DAFormer} is the first method applying transformer (SegFormer) on cross-domain natural scene image semantic segmentation task. Similar to our method, DAFormer also uses mmsegmentation to implement their method, so we can easily apply DAFormer on cross-domain RS image semantic segmentation tasks.
\par \textbf{Comparison experiments from Potsdam IR-R-G to Vaihingen IR-R-G. }In this task, Postsdam IR-R-G and Vaihingen IR-R-G images are respectively served as source-domain and target-domain. The 2904 annotated training images from Potsdam and 1296 no-annotation training images from Vaihingen are used to train the model. The 440 Vaihingen testing images are used for evaluation. The comparison results are shown in Tab.\ref{Tab1}. Compared to DeeplabV3 based methods, ST-DASegNet (DeeplabV3) surpasses the current SOTA method~\cite{DA-ResiDualGAN}. Compared to SegFormer based method (DAFormer~\cite{DAFormer}), ST-DASegNet (SegFormer) achieves a 2.03\% improvement on $mIoU$ value and 3.29\% improvement on $mF$-score. From Tab.\ref{Tab1}, we also find that even though the baseline model performs strong, ST-DASegNet (SegFormer) can still gain 20.22\% improvement on $mIoU$ value and 17.23\% improvement on $mF$-score. Particularly, on ``Clutter'' category, ST-DASegNet (SegFormer) shows outstanding performance over previous methods. On this single category, ST-DASegNet surpasses the second best method (DAFormer) by 18.77\% on $IoU$ value and 20.11\% on $F1$-score. 
\par \textbf{Comparison experiments from Vaihingen IR-R-G to Potsdam IR-R-G. }In this task, Vaihingen IR-R-G and Potsdam IR-R-G images are respectively served as source-domain and target-domain. The 1296 annotated training images from Vaihingen and 2904 no-annotation training images from Potsdam are used to train the model. The 1694 Potsdam testing images are used for evaluation. As shown in Tab.\ref{Tab2}, DeeplabV3 based ST-DASegNet performs much stronger than previous methods. Compared to previous SOTA method~\cite{DA-CIAUDA}, it achieves 3.30\% improvement on $mIoU$ value and 1.15\% on $mF$-score. SegFormer based ST-DASegNet also shows obvious superiority over DAFormer.
\begin{table*}	
	\centering
	\caption{Cross-domain RS image semantic segmentation comparison results (\%) from Potsdam R-G-B to Vaihingen IR-R-G. Methods with ``*'' are our reimplemented version.}
    \scalebox{0.7}{
    \begin{tabular}{ccccccccccccccc}
		\cmidrule(r){1-15}
  \multirow{2}{*}{Methods} &  \multicolumn{2}{c}{Clutter} & \multicolumn{2}{c}{Impervious surfaces} & \multicolumn{2}{c}{Car} & \multicolumn{2}{c}{Tree} & \multicolumn{2}{c}{Low vegetation} & \multicolumn{2}{c}{Building} & \multicolumn{2}{c}{Overall}
  \\ \cmidrule(r){2-15}
  {} & $IoU$ & $F1$-score & $IoU$ & $F1$-score & $IoU$ & $F1$-score & $IoU$ & $F1$-score & $IoU$ & $F1$-score & $IoU$ & $F1$-score & $mIoU$ & $mF$-score 
  \\ \cmidrule(r){1-15}
  DeeplabV3~\cite{DeeplabV3} (Baseline) & 0.58 & 1.16 & 40.42 & 57.57 & 12.52 & 22.25 & 30.88 & 47.19 & 12.12 & 21.62	& 54.23 & 70.33 & 25.12 & 36.68
  \\
  SegFormer~\cite{SegFormer} (Baseline) & 1.43 & 2.81 & 51.34 & 67.85 & 37.97 & 55.04 & 52.62 & 68.96 & 5.18 & 9.85 & 73.18 & 84.51 & 36.95 & 48.17
  \\
  \cmidrule(r){1-15}
  AdaptSegNet~\cite{AdapSegNet} & 2.99 & 5.81 & 51.26 & 67.77 & 10.25 & 18.54 & 51.51 & 68.02 & 12.75 & 22.61 & 60.72 & 75.55 & 31.58 & 43.05 
  \\
  ProDA~\cite{ProDA} & 2.39 & 5.09 & 49.04 & 66.11 & 31.56 & 48.16 & 49.11 & 65.86 & 32.44 & 49.06 & 68.94 & 81.89 & 38.91 & 52.70
  \\
  DualGAN~\cite{DA-DualGAN} & 3.94 & 13.88 & 49.16 & 61.33 & 40.31 & 57.88 & 55.82 & 70.66 & 27.85 & 42.17 & 65.44 & 83.00 & 39.93 & 54.82
  \\
  Bai~\textit{et al.}~\cite{DA-CLAL} & 10.80 & 19.40 & 62.40 & 76.90 & 38.90 & 56.00 & 53.90 & 70.00 & 35.10 & 51.90 & 74.80 & 85.60 & 46.00 & 60.00
  \\
  Zhang~\textit{et al.}~\cite{DA-CSLG} & 12.38 & 21.55 & 64.47 & 77.76 & 43.43 & 60.05 & 52.83 & 69.62 & 38.37 & 55.94 & 76.87 & 86.95 & 48.06 & 61.98 
  \\
  ResiDualGAN~\cite{DA-ResiDualGAN} & 9.76 & 16.08 & 55.54 & 71.36 & 48.49 & 65.19 & 57.79 & 73.21 & 29.15 & 44.97 & 78.97 & 88.23 & 46.62 & 59.84
  \\ 
  Wang~\textit{et al.}~\cite{DA-FGUDA} & 12.61 & 22.39 & 73.80 & 84.92 & 43.24 & 60.38 & 44.41 & 61.50 & 43.27 & 60.40 & 83.76 & 91.16 & 50.18 & 63.46 
  \\
  CIA-UDA~\cite{DA-CIAUDA} & 13.50 & 23.78 & 62.63 & 77.02 & \textbf{52.28} & \textbf{68.66} & 63.43 & 77.62 & 33.31 & 49.97 & 79.71 & 88.71 & 50.81 & 64.29
  \\ 
  DAFormer~\cite{DAFormer}* & 22.57 & 33.72 & 67.44 & 79.65 & 45.60 & 60.13 & \textbf{66.27} & \textbf{80.41} & \textbf{40.49} & \textbf{54.93} & 81.34 & 90.07 & 53.95 & 66.49
  \\ \cmidrule(r){1-15}
  ST-DASegNet (DeeplabV3) & 20.53 & 33.74 & 62.60 & 76.39 & 47.32 & 64.30 & 61.71 & 74.89 & 29.72 & 44.43 & 75.58 & 86.13 & 49.58 & 63.31
  \\
  ST-DASegNet (SegFormer) & \textbf{36.03} & \textbf{50.64} & \textbf{68.36} & \textbf{81.28} & 43.15 & 60.28 & 64.65 & 78.31 & 34.69 & 47.08 & \textbf{84.09} & \textbf{91.33} & \textbf{55.16} & \textbf{68.15}
  \\ \cmidrule(r){1-15}
   \end{tabular}
 }
 \label{Tab3}
 \end{table*}
 \begin{table*}	
	\centering
	\caption{Cross-domain RS image semantic segmentation comparison results (\%) from Vaihingen IR-R-G to Potsdam R-G-B. Methods with ``*'' are our reimplemented version.}
    \scalebox{0.7}{
    \begin{tabular}{ccccccccccccccc}
		\cmidrule(r){1-15}
  \multirow{2}{*}{Methods} &  \multicolumn{2}{c}{Clutter} & \multicolumn{2}{c}{Impervious surfaces} & \multicolumn{2}{c}{Car} & \multicolumn{2}{c}{Tree} & \multicolumn{2}{c}{Low vegetation} & \multicolumn{2}{c}{Building} & \multicolumn{2}{c}{Overall}
  \\ \cmidrule(r){2-15}
  {} & $IoU$ & $F1$-score & $IoU$ & $F1$-score & $IoU$ & $F1$-score & $IoU$ & $F1$-score & $IoU$ & $F1$-score & $IoU$ & $F1$-score & $mIoU$ & $mF$-score 
  \\ \cmidrule(r){1-15}
  DeeplabV3~\cite{DeeplabV3} (Baseline) & 4.61 & 8.82 & 46.02 & 63.03 & 59.71 & 74.77 & 1.63 & 3.53 & 7.1 & 13.25 & 37.34 & 54.37 & 26.07 & 36.30
  \\
  SegFormer~\cite{SegFormer} (Baseline) & 2.36 & 4.61 & 62.45 & 76.89 & 72.16 & 83.83 & 5.38 & 10.21 & 31.52 & 48.65 & 72.61 & 84.13 & 41.08 & 51.39
  \\
  \cmidrule(r){1-15}
  AdaptSegNet~\cite{AdapSegNet} & 6.11 & 11.50 & 37.66 & 59.55 & 42.31 & 55.95 & 30.71 & 45.51 & 15.10 & 25.81 & 54.25 & 70.31 & 31.02 & 44.75
  \\
  ProDA~\cite{ProDA} & 11.13 & 20.51 & 44.77 & 62.03 & 41.21 & 59.27 & 30.56 & 46.91 & 35.84 & 52.75 & 46.37 & 63.06 & 34.98 & 50.76
  \\
  DualGAN~\cite{DA-DualGAN} & \textbf{13.56} & \textbf{23.84} & 45.96 & 62.97 & 39.71 & 56.84 & 25.80 & 40.97 & 41.73 & 58.87 & 59.01 & 74.22 & 37.63 & 52.95
  \\
  Zhang~\textit{et al.}~\cite{DA-CSLG} & 13.27 & 23.43 & 57.65 & 73.14 & 56.99 & 72.27 & 35.87 & 52.80 & 29.77 & 45.88 & 65.44 & 79.11 & 43.17 & 57.77 
  \\
  Wang~\textit{et al.}~\cite{DA-FGUDA} & 10.84 & 17.49 & 66.11 & 79.75 & 65.45 & 80.17 & 28.64 & 43.51 & 35.47 & 51.85 & 68.63 & 81.32 & 45.86 & 59.74 
  \\
  CIA-UDA~\cite{DA-CIAUDA} & 9.20 & 16.86 & 53.39 & 69.61 & 63.36 & 77.57 & 44.90 & 61.97 & 43.96 & 61.07 & 70.48 & 82.68 & 47.55 & 61.63
  \\ 
  DAFormer~\cite{DAFormer}* & 1.07 & 1.88	& 65.12 & 78.16 & 70.40 & 84.28 &	\textbf{61.25} & \textbf{76.59} & 49.02 & 65.51 & 82.44 & 89.70	& 54.88 & 66.02
  \\ \cmidrule(r){1-15}
  ST-DASegNet (DeeplabV3) & 2.66 & 4.29 & 65.48 & 79.27 & 75.15 & 85.86 & 34.46 & 47.95 & 45.59 & 63.13 & 78.06 & 87.66 & 50.23 & 61.36
  \\
  ST-DASegNet (SegFormer) & 3.70 & 7.38 & \textbf{69.83} & \textbf{83.12} & \textbf{75.99} & \textbf{87.89} & 57.41 & 73.47 & \textbf{50.76} & \textbf{67.64} & \textbf{83.46} & \textbf{90.67} & \textbf{56.86} & \textbf{68.37}
  \\ \cmidrule(r){1-15}
   \end{tabular}
 }
 \label{Tab4}
 \end{table*}
\par \textbf{Comparison experiments from Potsdam R-G-B to Vaihingen IR-R-G. } In this task, Postsdam R-G-B and Vaihingen IR-R-G images are respectively served as source-domain and target-domain. The 2904 annotated training images from Potsdam and 1296 no-annotation training images from Vaihingen are used to train the model. The 440 Vaihingen testing images are used for evaluation. As shown in Fig.~\ref{Fig1}, ``Potsdam R-G-B to Vaihingen IR-R-G'' adaptation is more complex than ``Potsdam IR-R-G to Vaihingen IR-R-G'' adaptation. Besides suffering from different ground sampling distances, this task also suffers from remote sensing sensor variation. The comparison results on this task are shown in Tab.\ref{Tab3}. Obviously, SegFormer based ST-DASegNet outperforms DAFormer~\cite{DAFormer} and achieves SOTA results. DeeplabV3 based ST-DASegNet achieves comparable results with \cite{DA-CIAUDA}. In addition, Similar to the results on ``Potsdam IR-R-G to Vaihingen IR-R-G'' task (Tab.\ref{Tab1}), ST-DASegNet also outperforms the second best method by a large margin in ``Clutter'' category.
\par \textbf{Comparison experiments from Vaihingen IR-R-G to Potsdam R-G-B. }In this task, Vaihingen IR-R-G and Potsdam R-G-B images are respectively served as source-domain and target-domain. The 1296 annotated training images from Vaihingen and 2904 no-annotation training images from Potsdam are used to train the model. The 1694 Potsdam testing images are used for evaluation. From Tab.\ref{Tab4}, we also obtain encouraging results. On this task, DeeplabV3 based ST-DASegNet shows surprising improvement. Compared to baseline results, it gains 24.16\% on $mIoU$ value and 25.06\% and $mF$-score. Compared to the recent best DeeplabV3 based method~\cite{DA-CIAUDA}, it leads by 2.68\% on $mIoU$ value. Moreover, SegFormer based ST-DASegNet achieves the best performance on 4 categories (Impervious surfaces, Cars, Low vegetation, and Building) and achieves the current SOTA results on $mIoU$ value and $mF$-score.
\begin{table*}	
	\centering
	\caption{Cross-domain RS image semantic segmentation comparison results (\%) from Rural to Urban of LoveDA test dataset. Methods with ``*'' are our reimplemented version. ``AT'' indicates adversarial training and ``ST'' indicates self-training.}
    \scalebox{0.95}{
    \begin{tabular}{cccccccccc}
		\cmidrule(r){1-10}
  \multirow{2}{*}{Methods} & \multirow{2}{*}{Type} & \multicolumn{7}{c}{$IoU$} & \multirow{2}{*}{$mIoU$} 
  \\ \cmidrule(r){3-9} 
  {} & {} & Background & Building & Road & Water & Barren & Forest & Agriculture
  \\ \cmidrule(r){1-10} 
  DeeplabV3~\cite{DeeplabV3} (Baseline) & - & 32.81 & 43.41 & 27.59 & 79.23 & 14.84 & 29.24 & 20.97 & 35.44
  \\
  SegFormer~\cite{SegFormer} (Baseline) & - & 44.98 & 43.93 & 27.46 & 85.82 & 16.24 & 37.02 & 30.48 & 40.85
  \\ \cmidrule(r){1-10}
  DDC~\cite{DDC} & - & 43.60 & 15.37 & 11.98 & 79.07 & 14.13 & 33.08 & 23.47 & 31.53
  \\
  AdaptSegNet~\cite{AdapSegNet} & AT & 42.35 & 23.73 & 15.61 & 81.95 & 13.62 & 28.70 & 22.05 & 32.68
  \\
  FADA~\cite{FADA} & AT & 43.89 & 12.62 & 12.76 & 80.37 & 12.70 & 32.76 & 24.79 & 31.41
  \\
  CLAN~\cite{CLAN} & AT & 43.41 & 25.42 & 13.75 & 79.25 & 13.71 & 30.44 & 25.80 & 33.11
  \\
  TransNorm~\cite{Transform} & AT & 38.37 & 5.04 & 3.75 & 80.83 & 14.19 & 33.99 & 17.91 & 27.73
  \\
  PyCDA~\cite{PyCDA} & ST & 38.04 & 35.86 & 45.51 & 74.87 & 7.71 & 40.39 & 11.39 & 36.25
  \\
  CBST~\cite{CBST} & ST & 48.37 & 46.10 & 35.79 & 80.05 & 19.18 & 29.69 & 30.05 & 41.32
  \\
  IAST~\cite{IAST} & ST & 48.57 & 31.51 & 28.73 & 86.01 & \textbf{20.29} & 31.77 & 36.50 & 40.48
  \\
  DCA~\cite{DA-DeepCA} & ST & 45.82 & 49.60 & 51.65 & 80.88 & 16.70 & 42.93 & \textbf{36.92} & 46.36
  \\
  DAFormer~\cite{DAFormer}* & ST & 50.94 & \textbf{56.66} & \textbf{62.83} & \textbf{89.41} & 11.99 & 45.81 & 25.26 & 48.99
  \\ \cmidrule(r){1-10}
  ST-DASegNet (DeeplabV3) & AT+ST & 49.68 & 51.62 & 52.41 & 74.76 & 10.69 & 35.67 & 35.79 & 44.38
  \\
  ST-DASegNet (SegFormer) & AT+ST & \textbf{51.01}	& 54.23 & 60.52 & 87.31 & 15.18 & \textbf{47.43} & 36.26 & \textbf{50.28}
  \\ \cmidrule(r){1-10}
   \end{tabular}
 }
 \label{Tab5}
 \end{table*}
 \begin{table*}	
	\centering
	\caption{Cross-domain RS image semantic segmentation comparison results (\%) from Urban to Rural of LoveDA test dataset. Methods with ``*'' are our reimplemented version. ``AT'' indicates adversarial training and ``ST'' indicates self-training. }
    \scalebox{0.95}{
    \begin{tabular}{cccccccccc}
		\cmidrule(r){1-10}
  \multirow{2}{*}{Methods} & \multirow{2}{*}{Type} & \multicolumn{7}{c}{$IoU$} & \multirow{2}{*}{$mIoU$} 
  \\ \cmidrule(r){3-9} 
  {} & {} & Background & Building & Road & Water & Barren & Forest & Agriculture
  \\ \cmidrule(r){1-10} 
  DeeplabV3~\cite{DeeplabV3} (Baseline) & - & 26.64 & 48.14 & 23.37 & 43.97 & 11.35 & 32.47 & 50.61 & 33.79
  \\
  SegFormer~\cite{SegFormer} (Baseline) & - & 26.60 & 55.80 & 35.62 & 65.44 & 17.13 & 40.13 & 39.65 & 40.05
  \\ \cmidrule(r){1-10}
  DDC~\cite{DDC} & - & 25.61 & 44.27 & 31.28 & 44.78 & 13.74 & 33.83 & 25.98 & 31.36
  \\
  AdaptSegNet~\cite{AdapSegNet} & AT & 26.89 & 40.53 & 30.65 & 50.09 & 16.97 & 32.51 & 28.25 & 32.27
  \\
  FADA~\cite{FADA} & AT & 24.39 & 32.97 & 25.61 & 47.59 & 15.34 & 34.35 & 20.29 & 28.65
  \\
  CLAN~\cite{CLAN} & AT & 22.93 & 44.78 & 25.99 & 46.81 & 10.54 & 37.21 & 24.45 & 30.39
  \\
  TransNorm~\cite{Transform} & AT & 19.39 & 36.30 & 22.04 & 36.68 & 14.00 & 40.62 & 3.30 & 24.62
  \\
  PyCDA~\cite{PyCDA} & ST & 12.36 & 38.11 & 20.45 & 57.16 & 18.32 & 36.71 & 41.90 & 32.14
  \\
  CBST~\cite{CBST} & ST & 25.06 & 44.02 & 23.79 & 50.48 & 8.33 & 39.16 & 49.65 & 34.36
  \\
  IAST~\cite{IAST} & ST & 29.97 & 49.48 & 28.29 & 64.49 & 2.13 & 33.36 & 61.37 & 38.44
  \\
  DCA~\cite{DA-DeepCA} & ST & 36.38 & 55.89 & 40.46 & 62.03 & \textbf{22.01} & 38.92 & 60.52 & 45.17
  \\
  DAFormer~\cite{DAFormer}* & ST & \textbf{37.39} & 52.84 & 41.99 & 72.05 & 11.46 & 46.79 & 61.27 & 46.25
  \\ \cmidrule(r){1-10}
  ST-DASegNet (DeeplabV3) & AT+ST & 33.79 & 55.95 & 39.69 & 69.28 & 14.19 & 44.79 & 62.16 & 45.69
  \\
  ST-DASegNet (SegFormer) & AT+ST & 36.78 &\textbf{59.83} & \textbf{43.77} & \textbf{73.83} & 19.38 & \textbf{49.96} & \textbf{67.01} & \textbf{50.08}
  \\ \cmidrule(r){1-10}
   \end{tabular}
 }
 \label{Tab6}
 \end{table*}
 \subsubsection{Cross-domain RS image semantic segmentation on LoveDA}
\par Besides cross-domain adaptation between Potsdam and Vaihingen datasets, we further conduct comparison experiments on LoveDA dataset. Tab.\ref{Tab5} and Tab.\ref{Tab6} show the rural-to-urban and urban-to-rural results, respectively. From the results, we analyze the following aspects. (1) Compared to previous methods, our method provides insight into integrating self-training and adversarial training to tackle cross-domain remote sensing image semantic segmentation tasks. (2) Compared to baseline models, ST-DASegNet makes huge progress. On rural-to-urban adaptation task, ST-DASegNet (DeepLabV3) and ST-DASegNet (SegFormer) respectively surpass the baseline models by 8.94\% and 9.43\% on $mIoU$ value. On urban-to-rural adaptation task, ST-DASegNet (DeepLabV3) and ST-DASegNet (SegFormer) respectively surpass the baseline models by 11.90\% and 10.03\% on $mIoU$ value. (3) Compared to previous published SOTA methods, DCA~\cite{DA-DeepCA}, our proposed DeepLabV3 based ST-DASegNet achieves comparable results. It has 1.98\% inferiority on rural-to-urban adaptation task and 0.52\% superiority on urban-to-rural adaptation task. Our proposed SegFormer based ST-DASegNet outperforms DCA by a large margin on both these two tasks. (4) For a fair comparison with ST-DASegNet (SegFormer), we reimplement DAFormer. It is clear that ST-DASegNet (SegFormer) achieves better results. (5) These two tasks are online competitions. The track is ``LoveDA Unsupervised Domain Adaptation Challenge''. Among all the submitted results on rural-to-urban and urban-to-rural adaptation tasks, our method respectively ranks 4$^{th}$ and 2$^{nd}$ place. Among published methods, our method encouragingly achieves new SOTA results.
  \subsection{Ablation Study}
\par To separately show the performance of each component, we conduct ablation experiments on ``Potsdam IR-R-G to Vaihingen IR-R-G'' and ``Potsdam R-G-B to Vaihingen IR-R-G'' adaptation tasks shown in Tab.\ref{Tab7} and Tab.\ref{Tab8}. 
\subsubsection{Effectiveness of DDM} Before evaluating the performance of DDM, we first construct a dual path baseline model. This baseline model contains two student backbones and two student decoders. To form it, we simply remove DDM, discriminators and teacher network from ST-DASegNet. Compared to baseline model, it simply adds one more student network. During inference, two target predictions from the source and target decoders will be integrated with the soft voting strategy. Tab.\ref{Tab7} and Tab.\ref{Tab8} show that the dual path baseline model has limited improvement compared to the baseline model. This little improvement may come from the model ensemble strategy in the inference phase. 
\par Based on the dual path baseline model, we can easily insert DDM and evaluate its performance. As shown in Tab.\ref{Tab7} and Tab.\ref{Tab8}, when adding DDM, new models make huge progress. It means that the feature fusion and disentangling mechanism enhances the capability of source and target student backbones to extract different style features on different domain images. With DDM, the source and target student backbones can respectively represent the target images in source and target style with only source annotations. 
\begin{table*}	
	\centering
	\caption{Ablation comparison experiments on ``Potsdam IR-R-G to Vaihingen IR-R-G'' adaptation task (\%). ``Adv'' indicates adversarial learning. ``ST'' indicates self-training. }
    \scalebox{0.7}{
    \begin{tabular}{ccccccccccccccc}
		\cmidrule(r){1-15}
  \multirow{2}{*}{Methods} &  \multicolumn{2}{c}{Clutter} & \multicolumn{2}{c}{Impervious surfaces} & \multicolumn{2}{c}{Car} & \multicolumn{2}{c}{Tree} & \multicolumn{2}{c}{Low vegetation} & \multicolumn{2}{c}{Building} & \multicolumn{2}{c}{Overall}
  \\ \cmidrule(r){2-15}
  {} & $IoU$ & $F1$-score & $IoU$ & $F1$-score & $IoU$ & $F1$-score & $IoU$ & $F1$-score & $IoU$ & $F1$-score & $IoU$ & $F1$-score & $mIoU$ & $mF$-score 
  \\ \cmidrule(r){1-15}
  DeeplabV3~\cite{DeeplabV3} (Baseline) & 2.33 & 4.56 & 43.90 & 60.78 & 24.26 & 39.07 & 52.25 & 64.19 & 25.76 & 40.87 & 60.35 & 71.81 & 34.81 & 46.88 
  \\
  Dual path baseline model & 4.14 & 7.83 & 44.18 & 60.09 & 28.72 & 42.97 & 50.60 & 63.26 & 27.80 & 41.35 & 56.03 & 68.52 & 35.26 & 47.34
  \\
  + DDM & 6.02 & 13.72 & 58.10 & 72.67 & 18.62 & 33.28 & 54.90 & 70.04 & 38.80 & 56.08 & 66.05 & 77.70 & 40.42 & 53.92
  \\
  + DDM + Adv & 16.86 & 32.77 & 62.70 & 76.44 & 31.51 & 47.29 & 74.48 & 85.32 & 50.19 & 67.02 & 73.56 & 84.81 & 51.55 & 65.61
  \\
  + DDM + Adv + ST & 21.17 & 32.64 & 70.88 & 82.20 & 51.81 & 67.63 & 68.01 & 80.10 & 41.97 & 57.97 & 82.57 & 89.24 & 56.07 & 68.30
  \\
  with Single-target & 25.91 & 38.16 & 68.57 & 81.47 & 49.11 & 64.17 & 67.93 & 80.91 & 42.08 & 58.50 & 80.47 & 88.71 & 55.68 & 68.65
  \\ \cmidrule(r){1-15}
  SegFormer~\cite{SegFormer} (Baseline) & 4.22 & 9.47 & 61.03 & 76.07 & 31.13 & 47.89 & 66.31 & 78.87 & 44.47 & 60.38 & 75.50 & 87.95 & 46.11 & 60.11
  \\
  Dual path baseline model & 5.91 & 12.16 & 62.32 & 76.03 & 26.39 & 40.83 & 67.31 & 81.23 & 46.69 & 63.49 & 72.88 & 86.98 & 46.92 & 60.55
  \\
  + DDM & 27.53 & 44.01 & 61.61 & 73.46 & 36.98 & 54.05 & 63.46 & 74.70 & 47.17 & 62.75 & 79.74 & 88.73 & 52.75 & 66.28
  \\
  + DDM + Adv & 46.49 & 63.49 & 73.1 & 84.46 & 41.6 & 58.77 & 61.14 & 75.88 & 37.45 & 54.46 & 84.94 & 91.86 & 57.45 & 71.49
  \\
  + DDM + Adv + ST & 67.03 & 80.28 & 74.43 & 85.36 & 43.38 & 60.49 & 67.36 & 80.49 & 48.57 & 65.37 & 85.23 & 92.03 & 64.33 & 77.34
  \\
  with Single-target & 75.24 & 85.84 & 73.28 & 84.57 & 49.35 & 66.10 & 66.54 & 79.90 & 50.20 & 66.79 & 86.72 & 92.89 & 66.89 & 79.35
  \\ \cmidrule(r){1-15}
   \end{tabular}
 }
 \label{Tab7}
 \end{table*}
\begin{table*}	
	\centering
	\caption{Ablation comparison experiments on ``Potsdam R-G-B to Vaihingen IR-R-G'' adaptation task (\%). ``Adv'' indicates adversarial learning. ``ST'' indicates self-training.}
    \scalebox{0.7}{
    \begin{tabular}{ccccccccccccccc}
		\cmidrule(r){1-15}
  \multirow{2}{*}{Methods} &  \multicolumn{2}{c}{Clutter} & \multicolumn{2}{c}{Impervious surfaces} & \multicolumn{2}{c}{Car} & \multicolumn{2}{c}{Tree} & \multicolumn{2}{c}{Low vegetation} & \multicolumn{2}{c}{Building} & \multicolumn{2}{c}{Overall}
  \\ \cmidrule(r){2-15}
  {} & $IoU$ & $F1$-score & $IoU$ & $F1$-score & $IoU$ & $F1$-score & $IoU$ & $F1$-score & $IoU$ & $F1$-score & $IoU$ & $F1$-score & $mIoU$ & $mF$-score 
  \\ \cmidrule(r){1-15}
  DeeplabV3~\cite{DeeplabV3} (Baseline) & 0.58 & 1.16 & 40.42 & 57.57 & 12.52 & 22.25 & 30.88 & 47.19 & 12.12 & 21.62	& 54.23 & 70.33 & 25.12 & 36.68
  \\
  Dual path baseline model & 0.32 & 0.91 & 43.27 & 58.99 & 18.36 & 34.58 & 25.42 & 43.30 & 14.75 & 24.83 & 51.04 & 66.16 & 25.53 & 38.13
  \\
  + DDM & 2.80 & 5.44 & 50.72 & 67.30 & 18.42 & 31.11 & 54.34 & 70.42 & 21.34 & 35.18 & 50.35 & 66.98	& 32.99 & 46.07
  \\
  + DDM + Adv & 11.92 & 23.76 & 53.95 & 68.00 & 42.20 & 59.33 & 63.37 & 77.57 & 20.99 & 31.88 & 65.18 & 77.50 & 42.94 & 56.34
  \\
  + DDM + Adv + ST & 20.53 & 33.74 & 62.60 & 76.39 & 47.32 & 64.30 & 61.71 & 74.89 & 29.72 & 44.43 & 75.58 & 86.13 & 49.58 & 63.31
  \\
  with Single-target & 24.71 & 36.53 & 61.10 & 76.33 & 46.53 & 63.81 & 57.25 & 72.70 & 30.16 & 45.04 & 73.25 & 85.83 & 48.83 & 63.37
  \\ \cmidrule(r){1-15}
  SegFormer~\cite{SegFormer} (Baseline) & 1.43 & 2.81 & 51.34 & 67.85 & 37.97 & 55.04 & 52.62 & 68.96 & 5.18 & 9.85 & 73.18 & 84.51 & 36.95 & 48.17
  \\
  Dual path baseline model & 2.92 & 5.17 & 53.84 & 69.20 & 37.15 & 55.28 &	53.91 & 69.48 &	5.64 & 9.33 & 71.89	& 83.10 & 37.56 & 48.59
  \\
  + DDM & 13.42 & 27.26 & 55.22 & 71.13 & 37.33 & 54.37 & 53.71 & 69.86 & 10.65 & 19.20 & 78.86 & 88.18 & 41.53 & 55.00
  \\
  + DDM + Adv & 21.14 & 35.27 & 65.62 & 79.25 & 40.53 & 57.16 & 60.82 & 72.91 & 34.39 & 51.70 & 78.35 & 87.86 & 50.14 & 64.03
  \\
  + DDM + Adv + ST & 36.03 & 50.64 & 68.36 & 81.28 & 43.15 & 60.28 & 64.65 & 78.31 & 34.69 & 47.08 & 84.09 & 91.33 & 55.16 & 68.15
  \\
  with Single-target & 35.91 & 51.16 & 64.57 & 78.47 & 45.11 & 62.17 & 67.93 & 80.91 & 32.08 & 48.58 & 80.47 & 89.31	& 54.35 & 68.43
  \\ \cmidrule(r){1-15}
   \end{tabular}
 }
 \label{Tab8}
 \end{table*}
\subsubsection{Effectiveness of feature-level adversarial learning} After inserting DDM into the model, we further apply feature-level adversarial learning. From Tab.\ref{Tab7} and Tab.\ref{Tab8}, it is obvious that the performance of models is skyrocketing. DeepLabv3 based models improve by an average of 10.54\% on $mIoU$ value and 10.98\% on $mF$-score on ``Potsdam IR-R-G to Vaihingen IR-R-G'' and ``Potsdam R-R-B to Vaihingen IR-R-G'' adaptation tasks. SegFormer based models improve by 6.66\% on $mIoU$ value and 7.12\% on $mF$-score on the two tasks. These results demonstrate that feature-level adversarial learning can ease the domain shift problem by aligning single-style features from different domain images. Moreover, these results also prove that applying feature-level adversarial learning can enhance the power of the DDM.
\subsubsection{Effectiveness of EMA-based cross-domain separated self-training mechanism} Based on DDM and adversarial learning, we add EMA-based cross-domain separated self-training mechanism into the model and form the final ST-DASegNet. As shown in Tab.\ref{Tab7} and Tab.~\ref{Tab8}, with our proposed self-training mechanism, the performance gets further improved. From these results, it is clear that the models benefit from high-quality pseudo-labels, which suppress the representation tendency on source annotated images. The results also show that our proposed self-training mechanism can efficiently incorporate DDM and adversarial learning.
\begin{figure*}
\begin{center}
   \includegraphics[width=0.93\linewidth]{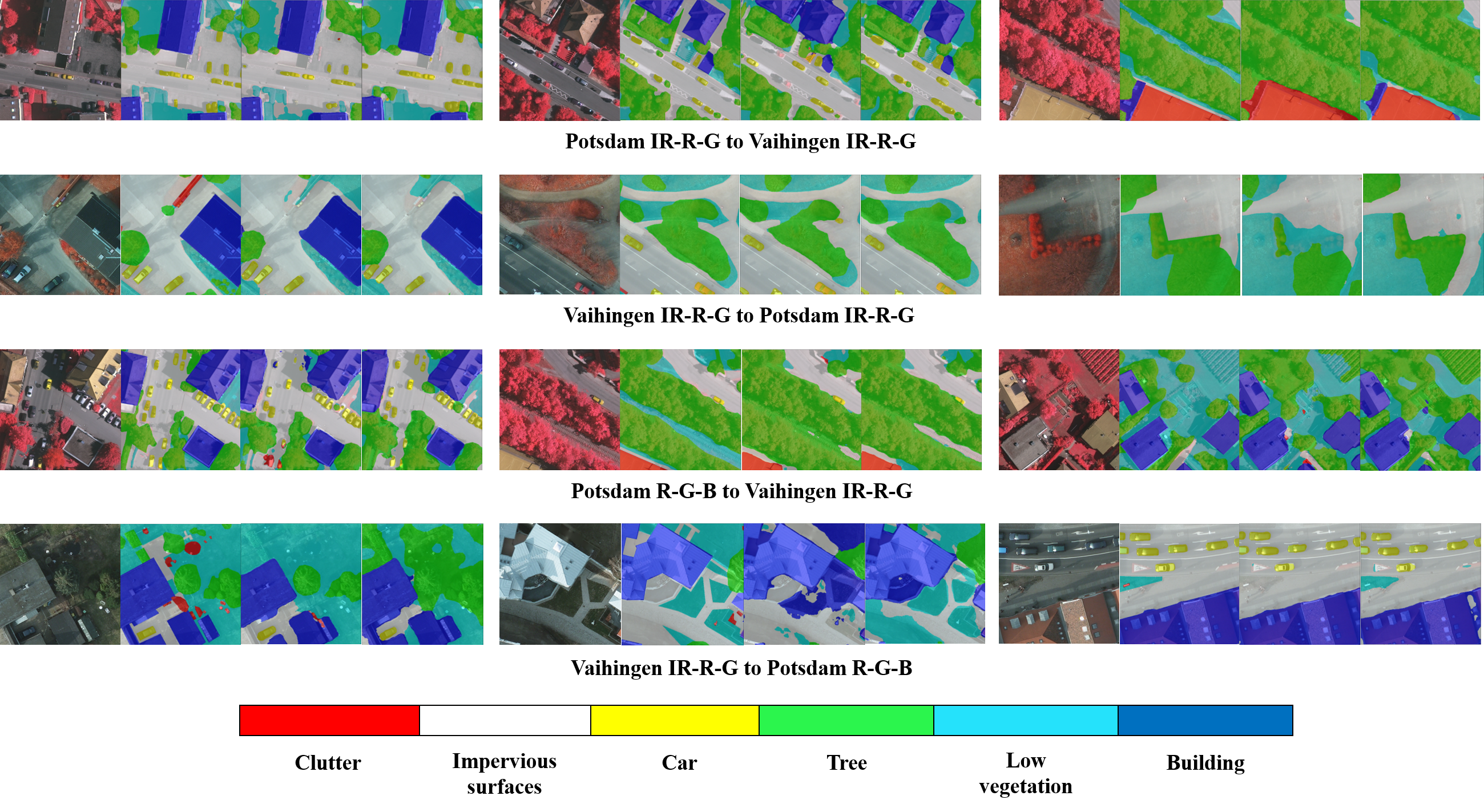}
\end{center}
   \caption{Qualitative visualization of results on Potsdam and Vaihingen datasets. For each task, we provide 3 cases containing 4 images. From left to right, the 4 images are respectively target images, ground truth, prediction of ST-DASegNet (DeeplabV3) and prediction of ST-DASegNet (SegFormer).}
\label{Fig5}
\end{figure*}
\begin{figure*}
\begin{center}
   \includegraphics[width=0.93\linewidth]{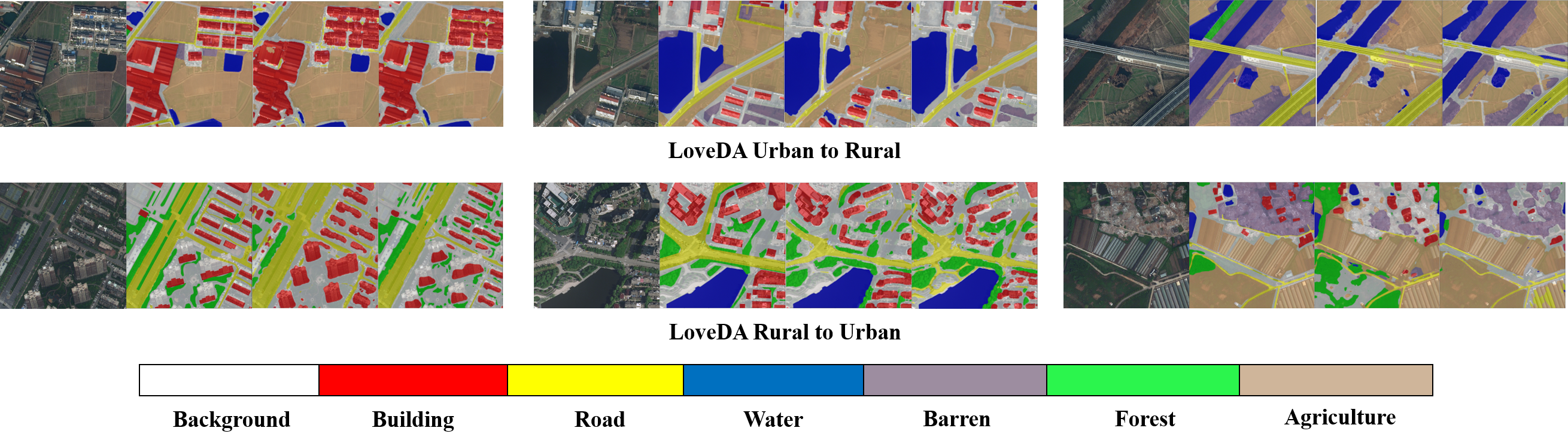}
\end{center}
   \caption{Qualitative visualization of results on LoveDA datasets. For each task, we provide 3 cases containing 4 images. From left to right, the 4 images are respectively target images, ground truth, prediction of ST-DASegNet (DeeplabV3), and prediction of ST-DASegNet (SegFormer).}
\label{Fig6}
\end{figure*}
\subsubsection{``Decoder-only'' Versus ``Single-target''} In this paper, we propose two self-training paradigms. To compare their performance, we conduct experiments in Tab.\ref{Tab7} and Tab.\ref{Tab8}. On ``Potsdam IR-R-G to Vaihingen IR-R-G'' task, ST-DASegNet (SegFormer) with ``Single-target'' paradigm outperforms ST-DASegNet (SegFormer) with ``Decoder-only'' paradigm by 2.56\% on $mIoU$ value and 2.01\% on $mF$-score. ST-DASegNet (DeepLabV3) achieves comparable results with these two paradigms. From the results of ``Potsdam R-G-B to Vaihingen IR-R-G'' task, ``Single-target'' paradigm has little inferiority compared to ``Decoder-only'' paradigm. Generally, ``Single-target'' and ``Decoder-only'' has close performance. Due to less training computation cost, we mainly adopt ``Decoder-only'' in this paper. Despite that ``Single-target'' can not beat ``Decoder-only'', it still provides another inspiring paradigm for EMA-based cross-domain separated self-training mechanism.
\par As a whole, DDM is designed to improve the model by enhancing feature representation capacity. Feature-level adversarial learning is used to ease the domain-shift problem. EMA-based cross-domain separated self-training mechanism makes an effect on balancing the representation tendency between source and target domain images. These three key components of ST-DASegNet show great compatibility, which improves the model from three different perspectives.
\begin{figure}
\begin{center}
   \includegraphics[width=1.0\linewidth]{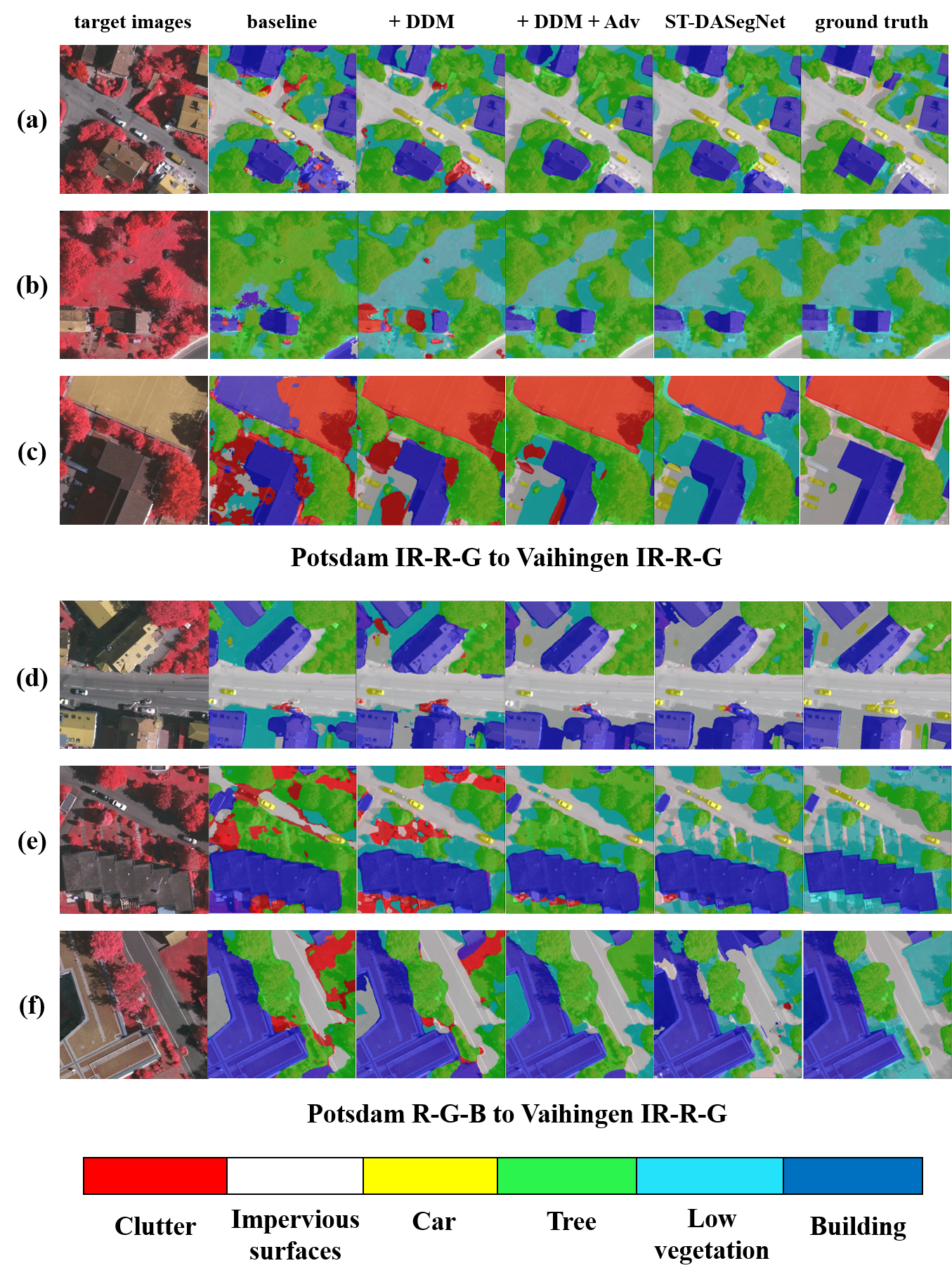}
\end{center}
   \caption{Ablation visualization on ``Potsdam IR-R-G to Vaihingen IR-R-G'' task (Tab.\ref{Tab7}) and ``Potsdam R-G-B to Vaihingen IR-R-G'' task (Tab.\ref{Tab8}). For each task, we provide 3 cases containing 6 images. From left to right, the first image is the target image. The rest 5 images are respectively predictions of the baseline model, model with DDM, model with DDM and adversarial learning mechanism, model with DDM, adversarial learning mechanism and self-training mechanism (ST-DASegNet), and ground truth. Here, we select SegFormer as the baseline model. Particularly, we do not provide results of the dual path baseline model, because it does not contain key components and improves little compared to baseline model.}
\label{Fig7}
\end{figure}
\subsection{Visualization and Analysis}
\subsubsection{Qualitative visualization of segmentation results} From Tab.\ref{Tab1} to Tab.\ref{Tab6}, we report the experimental results of our proposed ST-DASegNet on 6 cross-domain RS image semantic segmentation tasks. Here, we visualize the predictions using above-mentioned trained models to intuitively show the improvement of ST-DASegNet. 
\par As shown in Fig.~\ref{Fig5}, we can intuitively find that ST-DASegNet has strong performance on cross-domain RS image semantic segmentation tasks between Potsdam and Vaihingen datasets. Specifically, from the most right case on the ``Potsdam IR-R-G to Vaihingen IR-R-G'' task, we find ST-DASegNet has excellent perceptual ability on the ``Clutter'' category, which coincides with the results shown in Tab.\ref{Tab1}. From the middle and the most right cases on the ``Vaihingen IR-R-G to Potsdam IR-R-G'' task, ST-DASegNet shows a strong ability on distinguishing ``Tree'' and ``Low vegetation''. Similar results appear in Tab.\ref{Tab2}, where results of the ``Tree'' and ``Low vegetation'' categories improve by a large margin compared to baseline results. On the ``Potsdam R-G-B to Vaihingen IR-R-G'' task, the 3 visualization cases further confirm the results in Tab.\ref{Tab3}, where ST-DASegNet makes huge progress on segmenting ``Clutter'', ``Impervious surfaces'' and ``Building'' categories. On the ``Potsdam R-G-B to Vaihingen IR-R-G'' task, the most right case shows outstanding performance on recognizing ``Car''. As shown in Tab.\ref{Tab4}, both ST-DASegNet (DeeplabV3) and ST-DASegNet (SegFormer) surpass previous methods by a large margin on the results of the ``Car'' category.
\par As shown in Fig.~\ref{Fig6}, we provide the visualization results on LoveDA dataset. Since images in the testing dataset do not have annotations, we display the results of 
images in the validation dataset. It can be seen that ST-DASegNet can produce reasonable predictions. Especially on some hard categories like ``Barren'' and ``Forest'', ST-DASegNet can also predict well. The visualization results also coincide with the results in Tab.\ref{Tab5} and Tab.\ref{Tab6}.
\subsubsection{Visual ablation study}
\par To separately show the effectiveness of key components of ST-DASegNet in an intuitive manner, we provide ablation visualization on ``Potsdam IR-R-G to Vaihingen IR-R-G'' and ``Potsdam R-G-B to Vaihingen IR-R-G'' tasks shown in Fig.~\ref{Fig7}. When adding DDM, many large-region mistakes are avoided. In the baseline result of case (b), large regions of ``Low vegetation'' are classified into ``Tree''. In the baseline result of case (c), large regions of ``Clutter'' are classified into ``Building''. After adding DDM, these large-region mistakes are correctly classified. When adding adversarial learning, the edge of each category is clearer and the whole region of each category becomes smooth. In all 6 cases, we find that many small regions will be incorrectly classified into ``Clutter''. After adding adversarial learning, this problem is almost completely solved. When further applying EMA-based cross-domain separated self-training mechanism, many detailed mistakes for almost every category are corrected, which makes the results seem much better. As shown in cases (a), (b), and (f), some little rectifications on ``Building'' make predictions seem closer to ground truth. Similarly in cases (b) and (e), detailed corrections on ``Impervious surfaces'' really help improve the quality of predictions.
\par Theoretically, with DDM, source, and target student backbones can respectively represent the target images into source style and target style with only source annotations. In other words, besides domain universal features, some target-specific features can be extracted. However, feature inconsistency caused by domain shift will harm the effectiveness of DDM. When further adding feature-level adversarial learning, the domain shift problem is eased and cross-domain single-style features are aligned to become more consistent. Therefore, the feature representation capability of backbones and DDM are both improved. When integrating our proposed self-training mechanism, it can fundamentally improve this task by providing high-quality target annotations (pseudo-labels). In summary, besides ablation comparison experiments (Tab.\ref{Tab7} and Tab.\ref{Tab8}), ablation visualization further proves the effectiveness and great compatibility of the three key components of ST-DASegNet.  
\begin{figure*}
\begin{center}
   \includegraphics[width=0.92\linewidth]{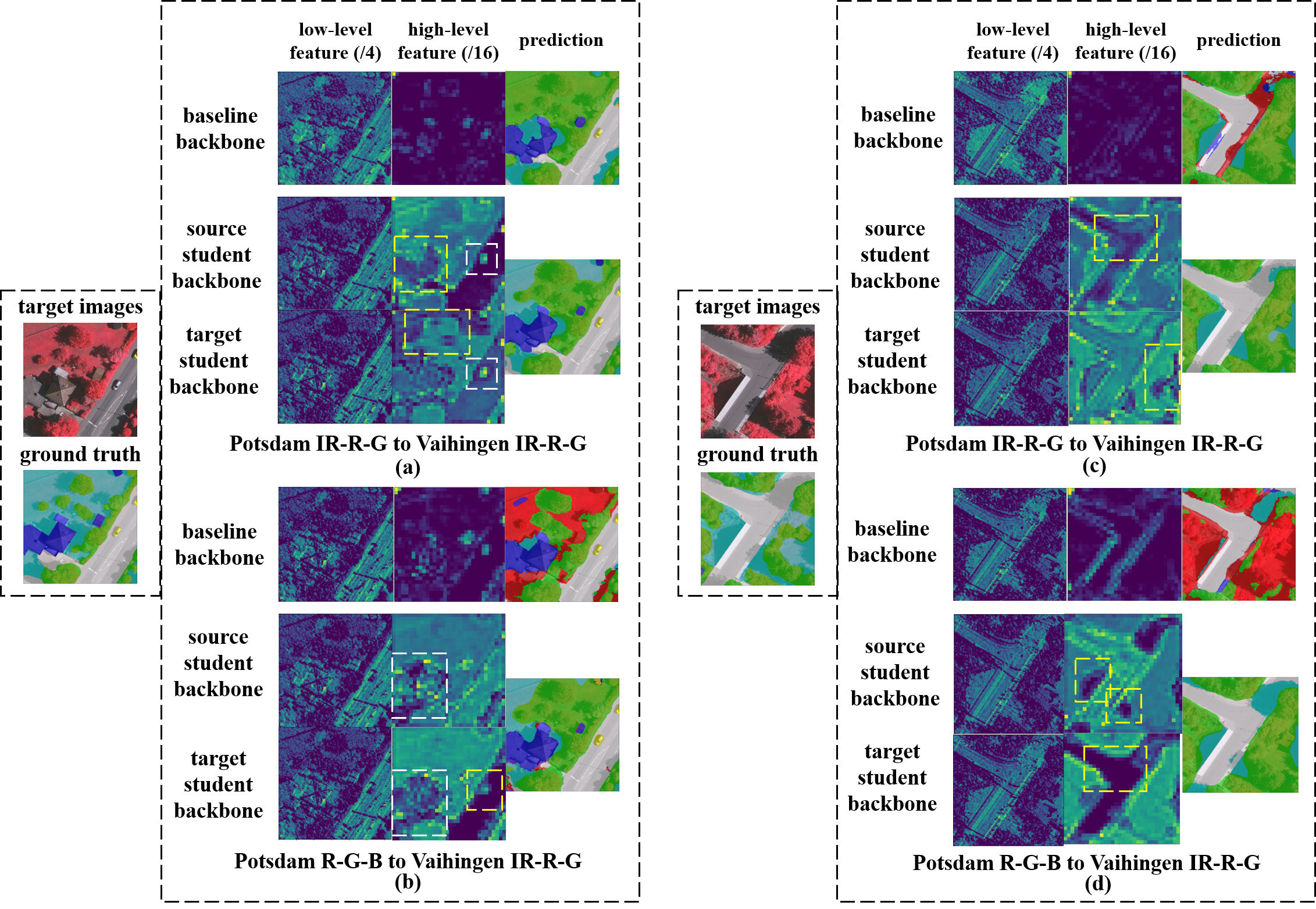}
\end{center}
   \caption{Feature map visualization comparison between baseline model and ST-DASegNet. We select the feature maps extracted from backbones for visualization. Feature maps listed here are channel-wise average feature maps. ``/4'' and ``/16'' respectively indicate the downsampling rate compared to the original image. ``Prediction'' indicates the final segmentation results of the baseline model and ST-DASegNet.}
\label{Fig8}
\end{figure*}
\subsubsection{Visualization and analysis on feature maps}
\par To intuitively show the interpretability of ST-DASegNet, we visualize the multi-level feature maps extracted from backbones and make an analysis. We select ``Potsdam IR-R-G to Vaihingen IR-R-G'' and ``Postdam R-G-B to Vaihingen IR-R-G'' tasks to conduct experiments. We select SegFormer as the baseline model.
\par \textbf{Feature map visualization comparison between baseline model and ST-DASegNet. } As shown in Fig.~\ref{Fig8}, we will analyze the visualization results with 4 cases from the following 2 points. (1) Baseline model and ST-DASegNet both have well-represented low-level feature maps, which have abundant detailed information and clear edges among regions of each category. (2) On high-level feature maps, it is obvious that two backbones of ST-DASegNet have better representation capability than  baseline backbones. As shown in the ``yellow box'' regions of the case (a), ST-DASegNet's high-level feature maps have clearer boundaries between regions of ``Tree'' and ``Low vegetation''. As shown in case (a) and case (b), we find that the edge of ``Building'' on ST-DASegNet's high-level feature maps is also in better shape. Case (c) and case (d) reveal that ST-DASegNet has better performance on localizing ``Low vegetation''. From Fig.~\ref{Fig8}, we find that the high-level feature maps of the baseline backbone will miss some important information. Even though the baseline backbone can represent well on some categories (e.g. Impervious surfaces in case (d)), the generalized representation ability is far behind ST-DASegNet. 
\par \textbf{Feature map visualization comparison between source and target student backbones of ST-DASegNet. }Cases in Fig.~\ref{Fig8} intuitively prove the effectiveness of our dual backbones structure. First, the two backbones of ST-DASegNet are complementary in representing target images. As shown in the ``yellow box'' regions of the case (a), the source student backbone can better find the ``Building'' edges while the target student backbone can better distinguish the boundaries between regions of ``Tree'' and ``Low vegetation''. In case (b), when the source student backbone fails to localize ``Car'', target student backbone successfully finds it. In cases (c) and (d), one backbone has advantages in dividing the edge of ``Impervious surfaces'' and the other has better performance on recognizing ``Low vegetation''. Second, two backbones can extract universal features. As shown in the ``white box'' regions of cases (a) and (b), the two backbones both have strong perceptions of  ``Car'' and ``Building''. Third, source and target student backbones in ST-DASegNet can respectively represent target images in source style and target style. In case (a), we find that the source student backbone tends to focus on ``Building'' and ``Impervious surfaces'', which is similar to the source-only trained baseline model. Target student backbone tends to have a better focus on ``Low vegetation'' because intuitively distinguishing ``Low vegetation'' and ``Tree'' in target images (Vaihingen IR-R-G) is easier than source images (Potsdam IR-R-G). From another perspective, if comparing case (c) with case (d), we find the representation style (preference) of the source style backbone completely changes, because source-domain images change from Potsdam IR-R-G to Potsdam R-G-B. 
\begin{figure*}
\begin{center}
   \includegraphics[width=0.95\linewidth]{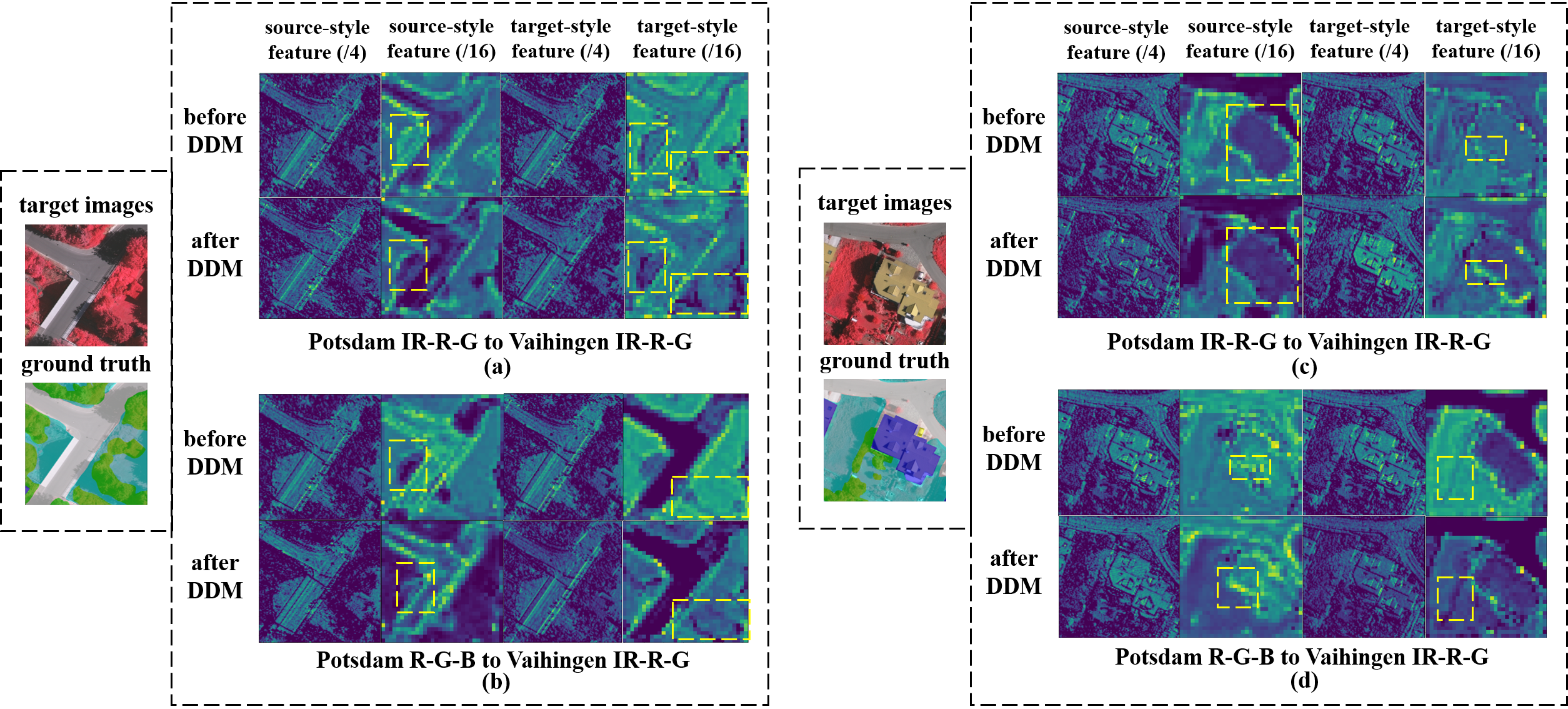}
\end{center}
   \caption{Visualization on domain disentangling. We select the feature maps extracted from backbones for visualization. Feature maps listed here are channel-wise average feature maps. ``/4'' and ``/16'' respectively indicate the downsampling rate compared to the original image.}
\label{Fig9}
\end{figure*}
\par \textbf{Visualization on domain disentangling. }Achieving high-quality domain disentangling with DDM is important for ST-DASegNet. With DDM, we expect to extract the invariant feature and purify the unique feature of source-style and target-style features. If DDM works well, more distinct target-style feature extraction can make an effect on better-representing target images. Moreover, differences between source-style and target-style features can improve the representation diversity. In Fig.~\ref{Fig9}, we adopt feature map visualization comparison to intuitively analyze domain disentangling. First, low-level features have minor changes because low-level features are always invariant, similar, and ``no-style''. Second, after DDM, high-level source-style and target-style features both become more distinct. In case (a) and (c), we observe that high-level source-style feature tends to focus on ``Impervious surfaces'' and ``Building'', which is also proved in Fig.~\ref{Fig8}. After DDM, the boundary of ``Impervious surfaces'' becomes clearer and expanded (case (a)). The expanded part is shadow and source student backbone tends to recognize it as ``Impervious surfaces''. In addition, after DDM, the boundary of ``Building'' is in better shape (case (c)). As a complementary, target-style feature tends to focus on ``Low vegetation'' and ``Tree''. The features after DDM obviously have more tendency in these two categories (case (a) and (c)). Particularly on the shadow area in case (a), the target student backbone tends to recognize it as ``Low vegetation''. Case (b) and case (d) also show that high-level source-style and target-style features become more distinct and different from each other, which proves the effectiveness of our proposed domain disentangling mechanism.
\section{Conclusion} 
\par In this paper, we propose a self-training guided disentangled adaptation method (ST-DASegNet) to tackle cross-domain remote sensing image semantic segmentation. In ST-DASegNet, we propose a dual path structure with two student backbones and two student decoders, which aims to represent source and target images into source style and target style with only source annotations. Based on this structure, we propose feature-level adversarial learning on cross-domain single-style features to enhance the representation consistency by feature alignment mechanism. Furthermore, we propose a domain disentangled module (DDM) on single-domain cross-style features to make source-style and target-style features more distinct and diverse. Besides adversarial learning and DDM, we propose and integrate an EMA-based cross-domain separated self-training paradigm (``Decoder-only'' and ``Single-target''). Our proposed self-training mechanism can balance the representation tendency between source and target domains by generating pseudo-labels of target images. The three key components show great compatibility, which can improve the performance of ST-DASegNet from three different perspectives. Extensive experiments on different benchmark datasets show that ST-DASegNet outperforms the previous SOTA methods on cross-domain RS image semantic segmentation tasks. Abundant visualization analysis intuitively proves that our proposed ST-DASegNet is reasonable and interpretable.

\section*{Acknowledgment}
This work was supported by the National Natural Science Foundation of China (grant numbers 62072021 and 61901015).
\bibliographystyle{cas-model2-names}

\bibliography{cas-refs}

\end{document}